\documentclass[conference]{IEEEtran}

\usepackage[T1]{fontenc}
\usepackage{lmodern}
\usepackage{cite}
\usepackage{amsmath,amssymb}
\usepackage{graphicx}
\usepackage{booktabs}
\usepackage{url}
\usepackage[hidelinks]{hyperref}

\begin{document}

\title{\fontsize{17}{20}\selectfont \textit{Bag of Tricks or Bag of Myths?}\\
Reducing Modeling Complexity with Task Knowledge in Explainable Suicide Risk
Assessment}

\author{
\IEEEauthorblockN{Shlok Shelat\textsuperscript{1}\quad Shrey Salvi\textsuperscript{1}\quad
Souvik Roy\textsuperscript{2}\quad Manas Gaur\textsuperscript{3}\quad Amit Sheth\textsuperscript{1}}
\vspace{2pt}
\IEEEauthorblockA{\textsuperscript{1}Indian AI Research Organisation\quad
\textsuperscript{2}Ahmedabad University\quad
\textsuperscript{3}University of Maryland, Baltimore County\\
\vspace{2pt}
\small\texttt{shlokshelat31@gmail.com, shrey.salvi@iairo.ai,
souvik.roy@ahduni.edu.in, manas@umbc.edu, amit@iairo.ai}}
}

\maketitle

% =====================================================================
\begin{abstract}
\normalfont
Assessing suicide risk from social media text is a small-data, high-stakes
setting that asks for more than a severity prediction. A useful system must
also identify the text supporting that judgment and detect clinically relevant
risk and protective factors. Yet standard NLP methods, among them larger
models, synthetic data, reweighted losses, ensembling and threshold tuning, are
routinely carried into settings like this one without testing whether their
reported gains survive severe class imbalance, coupled outputs and limited
author-level data. This is precisely the regime in which a practitioner has the
least held-out data with which to check them.

We study 1,635 clinician-annotated posts and audit 31 pre-specified techniques
from 7 methodological families through roughly 300 controlled experiments on
author-disjoint partitions. The techniques span model scaling, synthetic data,
loss reweighting, ensembling, structured prediction, thresholding, and
LLM-based methods. Searching the clinical NLP, suicide-risk NLP, and
controlled-audit literature, we found no prior audit of this playbook in this
regime. We then use the audit findings to construct a task-grounded system for
3 outputs: 4-level suicide risk, supporting evidence spans, and 24 clinical
risk and protective factors. Rather than combining techniques indiscriminately,
the system retains only those components the experiments support, and
introduces dependencies only where the data show them to be useful.

Only 5 of the 31 comparisons produced reliable gains, and those findings
shaped the submitted system directly. We reformulate clinical factor
prediction as entailment between each post and its corresponding codebook
definition, then combine members through an architecturally diverse ensemble
with class-balanced training and score rescaling. Risk predictions condition a
7-model evidence tagger ensemble, evidence spans restrict where symbolic risk
rules can act, and a difficult risk class is routed to a separate model. The
factor predictor remains independent, because risk evidence carries no factor
signal beyond that of arbitrary text of equal length. We also corrected a
mismatch between the validation score distributions used to fit thresholds and
the ensemble scores produced at test time, a repair we call
\textit{deployment-consistent calibration}, which yielded the largest single
improvement to the finished factor system. The result is 0.8203 for risk,
0.7953 for evidence and 0.7045 macro-F1 for clinical factors, a composite of
0.7781, placing third among 53 teams. \textit{The transferable finding is not
that placement but the principle behind it: the components that earned their
place were the ones justified by knowledge or structure already present in the
task, and identifying that knowledge is what let us remove the rest.} We call this
way of choosing components \textit{task-conditioned technique selection}. The shipped system is therefore plainer than the playbook would have made
it, though not cheaper to run.
\end{abstract}


% =====================================================================
\section{Introduction}

\begin{figure*}[!t]
\centering
\includegraphics[width=\textwidth]{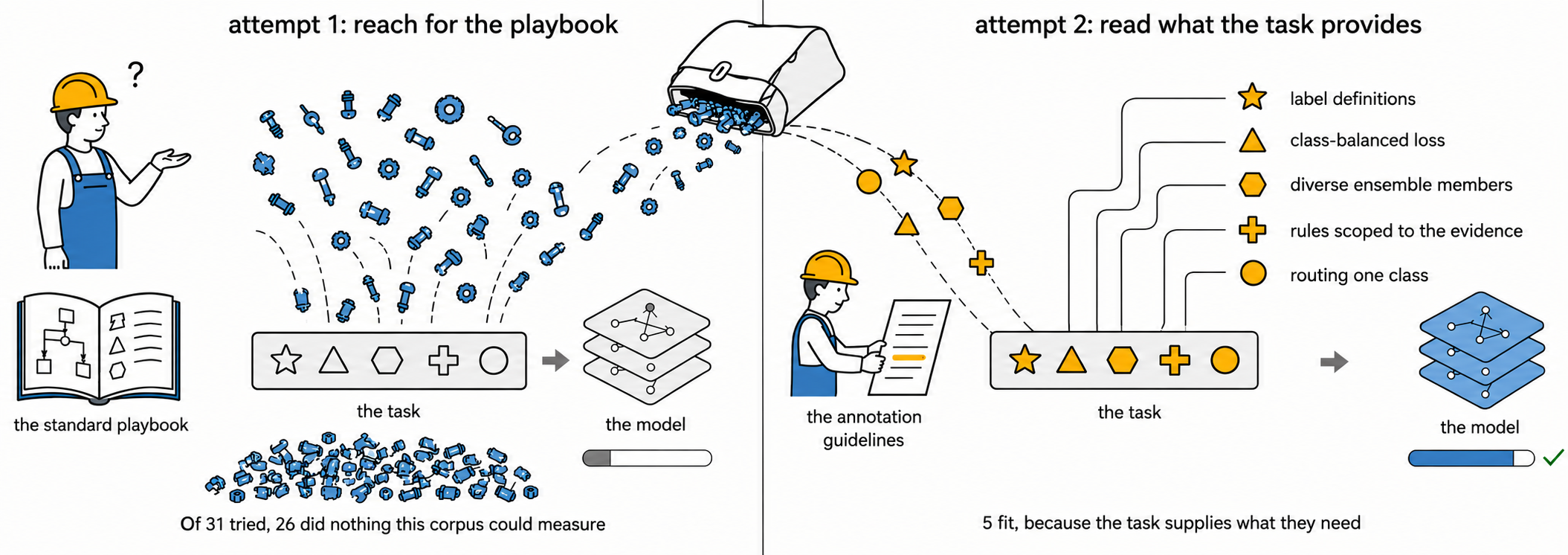}
\caption{\textbf{From a technique-driven playbook to knowledge-grounded
design.}\cite{sheth2021knowledge} Left: the standard response to a new task is to reach into the
playbook and add moving parts. We drew 31 such techniques, every one
recommended and individually well evidenced elsewhere; 26 produced no effect
this corpus can tell apart from the spread across repeated runs of one recipe.
Right: the alternative is to read what the task already provides and add only
what can exploit it. The 5 that held correspond one-to-one to 5 pieces of task
knowledge, and those divide into 2 kinds. Three are \emph{prior} knowledge,
readable before any model is trained: written label definitions motivate
definition-conditioned prediction, a steeply unbalanced label distribution
motivates class-balanced learning, and an annotator-marked justifying clause
motivates evidence-scoped rules. Two were \emph{discovered} only by experiment:
errors that differ across architectures motivate a diverse ensemble, and 1
class the main model is weakest on motivates selective routing. Both sets of
tools come out of the same bag; what separates them is the task, not where they
came from. \textit{Before adding modeling complexity, identify the knowledge or
structure available in the task and test whether a proposed component can
exploit it.}}
\label{fig:teaser}
\end{figure*}

More than 720,000 people die by suicide every year \cite{who2026suicide}.
In 50 years, research has not made it much easier to see who is at risk.
A meta-analysis of 365 studies found that longitudinal prediction of
suicidal thoughts and behaviors is only slightly better than chance, and
that this predictive ability has not improved over that whole period
\cite{franklin2017risk}. Part of the difficulty is where assessment
happens. It takes place in a clinic, with a person who has already walked
in. Many people never walk in. They do, however, write, and a great deal
of distress is written down on public online platforms long before anyone
intervenes. That is why there is now sustained interest in reading such
posts automatically and estimating risk from them
\cite{zirikly2019clpsych}. A risk score on its own, though, is not much use
to anyone. Whoever has to act on a flagged post needs to know how serious
it is, which words made it serious, and what in the writer's life is
raising or lowering the danger. The shared task studied here, a public
competition in which every entrant works from the same annotated posts,
asks for exactly those 3 things: a risk level on a 4-point scale, the
words that justify it, and which of 24 clinical risk and protective
factors the post shows. Each answer is scored between 0 and 1, and the 3 are combined into a single
composite number. The best entry this year reached
0.7951, and ours reached 0.7781, third of 53 on the public leaderboard at
the close of submission. The placement is not
what this paper is about. We used the shared task as a controlled test bed, because every system is
evaluated on the same annotated corpus under the same metric: 31
pre-specified comparisons drawn from the techniques our field reaches for by
default, each run one at a time under a single protocol. \textbf{Of those, 5 held and 26 did
not.} To our knowledge, no audit of this kind has been run on a task of this shape,
and that shape is what the argument rests on. But the audit is the evidence
rather than the point. \textbf{What it licenses is a way of designing: on a
corpus like this one, the knowledge the task already contains is what lets a
designer remove modeling choices rather than accumulate them.} Every
technique that survived did so because this task supplies the one thing that
technique needs, and knowing which 5 those were is what let us leave the
other 26 out of the system. Figure~\ref{fig:teaser} states that shift, and
the rest of the paper is the evidence for it.

What follows is less about suicide than about what happens when the
standard machinery of our field meets a problem shaped like this one.
Suppose you set out to build such a system. There is a familiar set of
moves you would reach for, and you would be in good company reaching for
them. You would start from a bigger pretrained model, one that has already
read enormous amounts of ordinary text before it ever sees your problem,
because bigger models are better at almost everything. You would gather
more labeled examples, and where you could not gather them, you would
generate some \cite{wei2019eda}. You would first let the model read text
from your own domain so it becomes accustomed to how people write there,
and only then train it on your labeled examples \cite{gururangan2020dapt}.
Rather than leave the model to infer each label from examples alone, you
would hand it a written description of the label and ask whether the text
matches \cite{yin2019entailment}. Because a handful of labels appear in
hundreds of posts while others appear in only 8, you would change the
loss so that the rare ones count for more \cite{lin2017focal,cui2019classbalanced}. Since no
single training run is entirely trustworthy, you would train several and
average them \cite{lakshminarayanan2017ensembles}. And because the model
hands back probabilities while the task wants decisions, you would tune a
cutoff for each label \cite{lipton2014thresholding}. Every one of these has
a paper behind it and a long record of working, and assembling a set of
them into a single recipe is itself a publishable contribution
\cite{he2019bagoftricks}. Call this the playbook. It is intuitive, it is
widely accepted, and we reached for it too.

But does the playbook hold on a problem shaped like this one? There is
reason to doubt it. A technique is
proposed, it is shown to work on the benchmark it was proposed on, and it
then travels into other people's pipelines on the strength of that one
result; it is rarely asked to prove itself again somewhere new. When somebody does stop and re-test a whole family of techniques
under one controlled protocol, much of the family fails. When dozens of
proposed improvements to the standard text model were re-tested separately
under one common setup, most gave no benefit \cite{narang2021transformer}.
The standard tricks for manufacturing extra training text stopped paying
off once modern pretrained models arrived \cite{longpre2020augmentation}.
Reweighting a training set to favor rare cases does considerably less than
practitioners assume \cite{byrd2019importance}, and of 18 published neural
recommendation systems only 7 could be reproduced at all, 6 of which were
beaten by simple baselines \cite{dacrema2019progress}.

None of those audits covered a task shaped like this one. They ran on
large benchmarks with one output per example and a metric that follows the
data's own frequencies. This task is the opposite on all 3 counts. What goes
unexamined is the combination: a corpus of 1,635 posts by 153 authors, 3
outputs that the annotation procedure makes dependent on one another, and a
macro-averaged score that gives a category with 8 positive examples the
same weight as one with 745.
These conditions create the design problem this paper is about: many
plausible modeling choices, and too little data to validate them
individually. That is an argument for reducing the number of moving parts by
exploiting what the task already tells you, rather than for accumulating
techniques and hoping. They are also the conditions under which shared tasks
in this area are actually decided. And to our knowledge, nobody has checked
the playbook in them. That absence matters for a practical reason
rather than a bibliographic one. \textit{A practitioner working here has the least data available for
testing a borrowed technique. That is exactly why they lean on the
published record instead. And that record has never been checked under
conditions like these.}

We expected the playbook to hold. Of the 31 comparisons we ran, 5 did, where
holding means the change won on a large majority of author-disjoint
resamples or cleared a margin no run-to-run variation we measured could
produce. Section~\ref{sec:protocol} makes that criterion exact before any
verdict is issued.
One early result changed how we read
everything that followed, and the rest of the paper repeats its pattern. The 24 factors come from a clinical
codebook, meaning the written manual the annotators worked from, so each
factor arrives with a definition in prose already attached. The model
therefore does not have to work out what hopelessness is from labeled examples
alone: we can hand it the definition and ask whether the post matches. Doing so lifted the factor score from 0.3795
to 0.6172, with the model, its size, and its training data all unchanged.
The same move ought to have helped with the words that justify a risk
level, and it did not. Given a written description of what counts as
evidence and asked which passages match it, a model reached 0.4830 against
0.7787 for an ordinary tagger, by which we mean a model that labels each
word in the post as inside or outside a justifying span. Those 2 gaps, 0.24 and 0.30, are far
larger than the 0.095 run-to-run spread we quantify below for the
factor component and larger than any spread we have measured anywhere, so
neither is an accident of measurement.

Why did the same move lift one output and sink the other? A factor label
has a definition that a post can be matched against. The words that justify
a risk level have none. The codebook never says which words are the right ones.
There is only an annotator's judgment about one particular post, so the
only description we could supply was the generic instruction to highlight
whatever justifies the label. That told the model nothing it did not
already have. The technique did not fail because it is a weak technique.
It failed because half of what it needs, an external definition to match
against, does not exist for that part of the task. That is the pattern the
rest of the paper repeats. \textbf{What paid was what this task could feed. What did not was what
needed something this task does not have.} Feeding a technique is not the same as inventing one. A loss that reweights
labels by how rare they are is an entirely general method. It held here because this task
really does have the steep imbalance it corrects for.

We therefore tested the playbook instead of following it. Across roughly
300 controlled experiments, spanning 31 pre-specified comparisons in 7
families of technique, we changed one variable at a time. Every offline
result was scored on partitions of the data that never place one author's
posts on both sides, and the few numbers read from the leaderboard instead are
marked as such wherever they appear. Each author often contributes several posts, and a careless
partition would let a model see a near-duplicate of the post it is being
scored on. Most of the tricks did not survive. Making the model
larger helped once, from the smallest size to the next, and the step after
that gave nothing. Generated training data never paid. A loss that concentrates
on the examples the model finds hardest outscored the ordinary one on a
single training run and lost on the next 2, where the only difference
between runs was the random seed. Five things held, and each of them exploited
something this task specifically provides.

The biggest single improvement to the finished system was none of these,
and it was not in the playbook at all. To choose the cutoff that turns a
probability into a decision, we had used models trained on parts of the
data. The system we actually submitted, which we call TRIDENT, averages 5 such
models inside each of its ensemble members, and averaging narrows the range
the scores occupy, so the
cutoff no longer falls where it was intended to. \textit{We had set the
line with one ruler and measured with another.} Putting it right was worth 0.0137 in factor macro-F1, measured out-of-fold
on the 1,635 annotated posts, and $+0.0041$ on the composite. It is the
largest increment anything made to the finished system, which is a different
quantity from the 0.2377 that definition conditioning gained against a naive
baseline, and it fit no new parameters. Our held-out protocol could not reveal it, because both the threshold
fitting and the comparison used single-fold held-out predictions rather than
the averaged distribution the system produces at test time.

We make 5 contributions, given in the order the argument builds.
\textit{First, we audit the standard playbook} on a task where, to our
knowledge, none of it had been re-tested: 31 pre-specified comparisons across 7
families of techniques, roughly 300 controlled runs, reported with the failures
given the same space as the successes. \textit{We are not aware of any audit of
comparable breadth in this regime; the individual techniques are well studied,
but they have not been re-tested together, under one protocol, on a task with
these properties.}

\textit{Second, we set out the standard of evidence that audit requires.} On a
corpus this small it is harder to meet than it looks. We re-run identical
recipes to measure how far apart repeated runs of one configuration land, and we
treat any effect smaller than that spread as noise rather than as a result
\cite{cawley2010selection,dodge2020finetuning}. Three runs of one recipe on the
factor component, differing only in the random seed, scored 0.5963, 0.6125 and
0.5176; \textit{that spread of 0.095 is wider than almost every effect we set
out to measure.}

\textit{Third, task-conditioned technique selection}
(Section~\ref{sec:calibration}): every technique needs some particular piece of
task knowledge, and on this corpus a technique paid exactly when the task
supplied it. That knowledge comes in 2 kinds, and the difference is practical: 3
of the 5 survivors rest on knowledge written into the annotation scheme before
any model is trained~\cite{gaurshethbook}, and could therefore have been chosen without spending any
compute, while the other 2 rest on measured properties of the data and of our
own models, which we could only find by looking. We did not start from this
rule; we arrived at it, and we test it on 4 matched pairs in which the same
intervention is run with that knowledge present and then absent, and either
changes sign or loses its effect.

\textit{Fourth, we use that evidence to simplify.} What the audit returns is a
subtraction rather than an extra layer: 9 of the 15 knowledge-to-component
chains in Table~\ref{tab:resource} terminate in nothing, and every one of those
is a component we implemented, measured, and took out again. What is left is the
submitted system, and for 4 of the 5 survivors we give a case where the same
technique fails, and the explanation says why.

\textit{Fifth, deployment-consistent calibration.} Thresholds tuned on the
predictions of single-fold models, then applied to an average of several, sit in
the wrong place: averaging narrows each member's score distribution, by a
different amount for each, which silently reweights the pool as well as
displacing the cutoff. Our offline protocol could not expose it \emph{by
construction}, because both sides of every comparison we ran inherited the same
mismatch; more generally, offline validation misses this whenever decisions are
calibrated on a prediction distribution the deployed system does not produce.
Correcting it fits no parameters and reads no labels, and it was worth more than
any modeling decision we made. \textit{Any system that averages fold models and
then thresholds them is exposed to the same defect, whatever the task.}

What we would most like carried forward is not the placement but the
principle behind it: on a corpus this size, the knowledge a task already
contains is enough to decide which components that task needs. A technique
earns its place only when the task supplies the thing it is built to
consume. \textit{Almost nothing in
the standard playbook does.}

% =====================================================================
\section{Related Work}

We connect 2 literature that are usually evaluated separately. The
first builds systems for this problem and takes the general-purpose
machinery for granted; the second questions that machinery and has
never turned to this problem.

Automatic assessment of suicide risk from social media has been studied
for close to a decade, and the field has converged on a small number of
carefully annotated resources rather than on scale. Several of them attach
clinically grounded severity labels to Reddit posts, in one case drawing
the scale directly from an instrument clinicians already use
\cite{gaur2019cssrs, gaur2018let, gaur2021characterization}. The CLPsych shared tasks have run on resources of
this kind since 2019. The organizers argued then that a binary at-risk flag
is not actionable, because it identifies more people than anyone has the
capacity to respond to, and that a graded assessment is needed if anyone is
to be prioritized \cite{zirikly2019clpsych, tsakalidis2022overview}. The 2024 edition went further and asked systems
to highlight the text supporting their judgments, because deciding
what to do about a flagged post is easier when a reviewer can see
why it was flagged \cite{chim2024clpsych}. The task studied here inherits
both moves and adds a third output, the 24 clinical risk and protective
factors introduced with the dataset the task is built on \cite{pfa2025}.
An earlier edition of this competition asked only for the risk level
\cite{cup2024overview}, and the strongest entry that year improved on
prior results by labeling the unlabeled test posts with its own model and
training on the result \cite{nguyen2024llm}. We tested that same technique
here and could not distinguish its effect from noise. It won a previous edition and did nothing here. Work in this area has also produced domain-adapted
encoders \cite{ji2022mentalbert} and knowledge-infused architectures \cite{8970629, sheth2020shades} that
build clinical structure into the model rather than leaving it to be
learned \cite{dalal2024cross,gaur2024crest, sinha2026llms}. Others have compared model families on
imbalanced clinical text \cite{mohammadi-etal-2024-welldunn, lu2022imbalance, roy2023process, sheth2022process}, and shown that a strong
pretrained encoder need not beat a simpler baseline on a long-tailed
clinical labeling task \cite{ji2021magicbert}. None of this work asks
whether the general-purpose techniques these systems are built from behave
here as they do elsewhere.

The second literature asks exactly that question of other fields, and its
results are consistent and unflattering. Re-tested together under one
protocol, most proposed improvements to a standard architecture give
nothing \cite{narang2021transformer}, and a properly tuned older model
matches the architectures that displaced it \cite{melis2018sota}. Training
procedure rather than model novelty accounts for much of the reported
progress in knowledge-graph embedding \cite{ruffinelli2020olddog}. A decade
of apparent progress in metric learning proves marginal once the
comparisons are made fairly \cite{musgrave2020metric}. In the small-data
regime specifically, few-shot methods re-evaluated under a corrected
protocol lose much of their reported advantage \cite{zheng2022fewnlu}, and
several proposed remedies for unstable fine-tuning turn out to be
misattributed \cite{zhang2021fewsample}. A related strand examines not the
techniques but the measurements that judge them. System rankings do not
survive a change in the data split \cite{gorman2019splits}. Tuning a model
on the same small sample used to score it inflates that score by about as
much as the differences between the methods being compared
\cite{cawley2010selection}. Typical experiments are too small to detect the
effects they report, and they exaggerate the ones they do detect
\cite{card2020power}. Changing only the random seed moves a score enough to
reverse a comparison \cite{reimers2017seeds}. Outside NLP,
the same exercise has repeatedly found that the reported progress of a
whole subfield rests on undertuned baselines
\cite{dacrema2019progress,oliver2018ssl}, and in clinical prediction from
structured data a review of 71 studies found no benefit from the machine
learning playbook over logistic regression \cite{christodoulou2019ml, alambo2019question}.

Audits of this kind re-implement a family of standard techniques, test
them one at a time under a single protocol, and report the failures
alongside the successes. They exist for architecture modifications \cite{narang2021transformer}, data augmentation
\cite{longpre2020augmentation}, semi-supervised learning
\cite{oliver2018ssl}, knowledge-graph embedding training
\cite{ruffinelli2020olddog}, few-shot understanding \cite{zheng2022fewnlu},
and for entire subfields outside NLP
\cite{melis2018sota,dacrema2019progress,musgrave2020metric}.
\textbf{None, to our knowledge, has been run on a small, multi-output
clinical text task whose score is a macro-average over steeply unequal
label counts}. That is the regime in which shared tasks of this
kind are actually decided, and the one in which several of these techniques
are most often recommended. The nearest work in our own venue compares 3
adaptation strategies for a large language model on single-label
mental-health classification scored by accuracy \cite{kermani2025clpsych};
the nearest in method establishes this style of audit for architecture
modifications in a large-data, single-output setting where a change is
measured against a near-saturated baseline \cite{narang2021transformer}.
We ask the same question one level up the stack, of the applied technique
playbook rather than the architecture, in the regime where these
techniques are actually reached for. Two nearby literatures do not close
the gap. Cross-system tables in shared-task overviews report what many
teams did, but they confound technique with team, model, budget and seed,
which a controlled protocol does not, and comparative-methods studies in
medical informatics compare model families on independent outcomes rather
than holding a model fixed and varying the technique
\cite{lu2022imbalance,christodoulou2019ml}. The absence is not an absence
of interest. Small annotated clinical corpora with several coupled outputs
and a metric that protects the rare category are the normal case in this
area rather than an edge case. \textit{The published record therefore carries more
weight here than anywhere else and has been checked here least.}

Finally, the techniques that did survive here are borrowed rather than
invented. Handing a model the written definition of a label, rather than
making it learn the label from examples alone, is an established method
\cite{yin2019entailment}. It is also known to help most when labeled data
is scarce: across 8 tasks with between 100 and 2,500 labeled texts, it
gained 10 to 18 points over classical models without transfer learning
\cite{laurer2024nli}. Our corpus of 1,635 posts
sits squarely in that range, so before we ran anything the literature
predicted that this would be the strongest lever available to us. That is
the rule of this paper working in the one direction in which it can be checked before the fact. The literature named the condition the technique needs,
which is scarce labeled data. We confirmed that this task meets it. The
technique held.

% =====================================================================
\section{What This Task Supplies, and What It Withholds}
\label{sec:task}

This task hands over 2 different kinds of thing, and the difference matters
for everything that follows. We call them both \emph{task knowledge}, and
separate them by when they become available. The first kind is written down
in the annotation scheme before any model is trained: a clinical codebook that
defines each of the 24 factors in prose, and an annotator-marked clause
behind every risk judgment. The second is \emph{measured properties} of the
data, which we had to go and find: the corpus is small and its posts are not
independent, the score weights a rare category exactly like a common one,
and 2 of the 3 outputs are bound together by the annotation procedure.
Sections \ref{sec:protocol} and \ref{sec:calibration} return to that split,
because a technique justified by the first kind can be chosen before
spending any compute, and one justified by the second cannot.

\textbf{One post in, three answers out.} Given one Reddit post, a system must return a
risk level, the words that justify it, and a set of clinical factors. The
risk level is 1 of 4 ordered values: \emph{indicator}, meaning the post
concerns suicide but shows no risk to the writer; \emph{ideation}, meaning
the writer expresses thoughts of suicide; \emph{behavior}, meaning the
writer describes a plan or an act of self-harm; and \emph{attempt},
meaning the writer describes a suicide attempt of their own. The evidence
is the annotators' answer to the question \emph{why did you say that}: the
exact words in the post that support the level they chose. It is quoted
rather than paraphrased, so a system's evidence must appear verbatim in
the post. Factor labels represent a different kind of judgment. Rather
than describing how acute the moment is, they describe the writer's
circumstances: whether the post shows hopelessness, an absence of people
to turn to, substance use, a recent loss, and so on. Of the 24, 19 name
something that raises risk and 5 name something that lowers it, such as a
sense of responsibility toward others or a stated reason for living. They
are read from the post in their own right rather than deduced from the
risk level, and a post may show several of them, or none
\cite{pfa2025}.

\textbf{The unit is the author, not the post.} We work from 1,635
annotated posts written by 153 authors, and the blind leaderboard set is a further 378 posts by 36 authors whose
labels we never see. We partition the
1,635 two ways, both by author: a fixed validation split of 372 posts by 32
authors, used for quick comparisons, and a 5-fold cross-validation over all
153 authors, used whenever a number had to be trusted. Every corpus size
quoted later in the paper is one of these.

Those splits are made by author for a reason. Posts are not independent:
each author contributes several,
and posts by the same person often resemble one another, so the effective
number of independent units is closer to the number of authors than to the
number of posts. The dependence is strongest exactly where it does most
damage. Knowing only that an author's other posts carry a given category, without
reading the post at all, raises the odds that this post carries it too: by
a factor of 6.2 for the rarest categories, and 1.1 for the most common. A split made by post rather than by author would hand a model that
advantage on precisely the rare categories the score weights most heavily.
\textbf{Every split we report is therefore made by author, never by post.} Table~\ref{tab:data} gives the distribution of risk levels and,
alongside it, how often the annotators marked any justifying words.

\begin{table}[t]
\centering
\caption{Risk levels in the 1,635 annotated posts, and how many carry
marked evidence. Every post without evidence sits at the lowest level, and
every post above that level carries evidence.}
\label{tab:data}
\begin{tabular}{lrrr}
\toprule
Risk level & Posts & With evidence & Without evidence \\
\midrule
indicator & 611  & 23   & 588 \\
ideation  & 519  & 519  & 0   \\
behavior  & 391  & 391  & 0   \\
attempt   & 114  & 114  & 0   \\
\midrule
total     & 1{,}635 & 1{,}047 & 588 \\
\bottomrule
\end{tabular}
\end{table}

\textbf{The 3 outputs are scored by 3 different rules.} Each output is scored separately, and
the 3 are combined as $0.4\,R + 0.3\,E + 0.3\,F$. The risk term $R$ is
a weighted F1 over the 4 levels, so a level that occurs often counts for
more than one that occurs rarely. The factor term $F$ is a macro-F1 over
the 24 categories, which is the opposite convention: every category
contributes one twenty-fourth of the total regardless of how often it
appears. The evidence term $E$ is computed per post. A predicted span
counts as matching a gold span when one contains the other and the
prediction is no more than 3 times the gold span's length in words,
and each gold span may be matched only once. Precision and recall are
formed from those matches and combined into an F1 for that post, and the
scores of all posts are then averaged. A post with no gold evidence scores a
perfect 1.0 if the system predicts nothing.

\textbf{The scoring rule shapes almost every result below.} It has 3 consequences. First,
because the factor term treats all 24 categories alike, a category with 8
positive examples in the corpus carries exactly the weight of one with
745, so a technique that quietly stops predicting a rare category pays for
it far out of proportion to how often that category occurs. Second,
because the evidence term averages over posts rather than over spans, one
post counts the same whether the annotators marked one phrase in it or
five, and \textit{the 588 posts with no marked evidence can be answered perfectly
by predicting nothing at all}. Third, the length-ratio rule means a
prediction is not punished for being somewhat longer than the phrase it
matches, but is scored as a miss the moment it exceeds that bound. Span
length is a decision, not a stylistic choice.

\textbf{The outputs are not independent.} Table~\ref{tab:data} shows the
relationship that governs the rest of this paper. The annotators worked in
one direction: they chose the risk level first and then marked the words
that supported it. The consequence is exact in one direction and strong in the other. Of the 588
posts with no marked evidence, every one sits at the lowest risk level, and of
the 1,024 posts above that level, every one carries evidence. Read the other
way it is tight but not absolute: 588 of the 611 lowest-level posts have an
empty evidence field and 23 do not. A system that places a post at the lowest
level and empties its evidence is therefore following a rule that holds for 96
percent of such posts, and when the risk decision itself is wrong it loses the
post's evidence score along with its risk score. Risk and evidence share an
annotation dependency strong enough that errors in the first propagate into
the second.

% =====================================================================
\section{Deciding Whether a Change Is Real}
\label{sec:protocol}

What can a corpus of 1,635 posts actually resolve? Less than we expected,
and the answer governs every verdict in this paper. We report 2 different
quantities and keep them apart throughout. The \emph{seed range} is how far apart repeated
runs of an identical recipe land in absolute score: 0.095 on the factor
component and 0.0183 on the risk component. The \emph{null band} is the
smallest paired difference the resampling instrument of
Table~\ref{tab:instrument} can resolve. On both components the seed range is
wider than almost every effect the rest of the paper measures, and that is
why most of our verdicts are negative. The same instrument
still resolved a gain of 0.2377 and a gain of 0.0084, so a negative verdict
here means a technique did nothing measurable, not that we were unable to
look.

\textbf{Why the obvious protocol does not work here.} The usual way to
judge a change is to hold out some data, score the old system and the new
one on it, and keep the change if the number goes up. That procedure needs
the held-out set to be large enough that the number means something, and
here it is not large enough. The 372 posts we began with come from only 32
authors, and because posts by the same person resemble one another, the
effective number of independent observations is nearer 32 than 372. The
standard error of a single factor score on that set is about 0.0215, which
is larger than most of the effects we wanted to detect. Pooling all 5
folds to 1,635 posts brings it to roughly 0.0030. Both figures are the
spread of the statistic across author-disjoint resamples rather than an
analytic standard error, so they do not scale as the square root of the
sample size.
Table~\ref{tab:instrument} gives both, together with the smallest paired
difference each instrument can resolve.

\begin{table}[t]
\centering
\caption{The 2 instruments we used. The null band is the range within which
a paired difference between 2 systems cannot be distinguished from chance.
We construct it by scoring 2 identical configurations, differing only in
seed, on 50 resamples drawn without replacement at the author level, each
holding out a disjoint set of authors; the band is the 95th percentile of
the absolute paired differences so obtained. Because the resamples are drawn
from 1 fixed author pool they overlap, which is why we read win rates off
this instrument rather than intervals around a mean.}
\label{tab:instrument}
\begin{tabular}{lrrrr}
\toprule
Instrument & Posts & Authors & Std.\ err. & Null band \\
\midrule
single split      & 372   & 32  & 0.0215 & $\pm 0.056$ \\
pooled 5 folds & 1{,}635 & 153 & 0.0030 & $\pm 0.014$ \\
\bottomrule
\end{tabular}
\end{table}

\textbf{Measuring the noise before measuring anything else.} Before
judging any technique we ran one recipe 3 times, changing nothing but the
random seed. The 3 runs scored 0.5963, 0.6125 and 0.5176. Nothing
distinguishes them; they are the same system.
\textit{The gap of 0.095 between the best and the worst is the instability
of the recipe itself.} Three runs is too few to estimate a distribution, and
we do not treat this as one: it is a descriptive diagnostic for comparisons
the paired instrument cannot reach, not a statistical test, and the max--min
range of a small sample understates the spread of a larger one. We use it in
the one direction that error permits, to withhold verdicts rather than to
grant them. A difference smaller than this observed range cannot be separated from seed
variation using these replications alone. The paired instrument described below resolves far smaller effects,
and most of what we accepted was accepted on it. We report this before anything else
because it determines what the rest of the paper is allowed to conclude,
and because most of the techniques we tested produced effects well inside
that range.

We repeated the exercise on the risk component, where 2 runs of one recipe
differing only in seed scored 0.7283 and 0.7100. That seed range of 0.0183 is
far narrower than the factor component's, which is what a 4-way
classification should look like beside a 24-way multi-label problem, and it
errs in a useful direction. The distance between the extremes of 2 runs is
on average smaller than the distance between the extremes of 3, so a range
read off 2 runs understates the one a longer series would give. Every risk result we dismissed as too small to see was
judged against a bar that is, if anything, too low.

\textbf{The null results are not saturation.} A null result means
one of 2 things: the technique did nothing, or there was
nothing left for it to recover. The corpus lets us separate them,
because it contains posts that are near-duplicates of one another. On the
163 pairs whose token overlap exceeds 0.9, so close that a reader would
struggle to tell them apart, the annotators assigned the same risk level
93.3 percent of the time and agreed on 98.7 percent of factor cells. Two
posts a reader cannot distinguish therefore receive different risk labels
about 1 time in 15, and no deterministic system can be right on
both. Our risk output reaches 0.8203 and our factor output 0.7045, both
well below what that disagreement rate permits, so saturation does not
explain the null results we report. The arithmetic will not carry further than
that. Near-duplicates are not a random sample of the corpus, so the figure
bounds the ceiling rather than locating it.

\textbf{Scoring 2 systems on the same posts reveals far more.} Comparing 2
absolute scores fails to exploit the fact that both systems saw the same
posts. We instead score both on the same resample and record the
difference, then repeat over many resamples drawn by author, so that no
author's posts ever appear on both sides of a split. The paired difference
is far steadier than either score alone, which is what makes small effects
measurable at all: on the pooled instrument, differences outside
$\pm 0.014$ can be resolved, whereas absolute scores on the single split
cannot resolve anything below about 0.056. \textit{The paired design is therefore more sensitive than a comparison of
absolute scores, though effects smaller than the null band remain unresolved
rather than refuted.}

\textbf{Why we report win rate rather than a confidence interval.} The
natural summary of many paired differences is a confidence interval around
their mean. Here it is the wrong one. The resamples are drawn from one
fixed set of authors, so they overlap heavily and are not independent
draws; the interval around their mean therefore narrows as we draw more
resamples, whether or not the effect is real. That lesson was expensive.
One change produced a mean gain whose interval excluded zero on this
instrument, and we submitted it; \textbf{the change lost 0.0261 on the
leaderboard}. We now report instead the fraction of resamples on which a
change wins, together with the spread across resamples, and accept a change only when it wins on at least 80 percent of them. That
threshold, and the 90 percent below it, are conventions we fixed before
running the comparisons rather than quantities we calibrated; we report the
win rate for every result so that a reader who prefers a different bar can
apply it. Only 3 of the 5 techniques we accepted were judged on this rule at all, and
their win rates are 89, 93 and 99 percent, so none of them sits near the
bar: no verdict in this paper turns on a shift of a few points in it. The
other 2 survivors cleared effect sizes far outside any band we measured, and
were accepted on that basis rather than on a win rate. The null band is
the smallest difference a \emph{single} paired comparison can resolve; it
is not a threshold on the mean. A gain of $+0.0084$ that wins on 89 percent
of resamples is a small effect established consistently. That is a
different claim from a large effect established once, and it is the first
that this instrument is built to detect. For a change chosen from a large set of
candidates we require 90 percent, for the reason given below.

\textbf{Picking the best of 8 candidates inflates its gain by 0.014.} When several
candidate changes are measured and the best is kept, the winner's score
carries the effect plus whatever noise happened to favor that candidate.
The size of that inflation follows from the spread of the paired
differences, which we measured at about 0.007 across 153 authors.
Selecting the best of 8 candidates therefore inflates the winner's apparent
gain by roughly $0.007\sqrt{2\ln 8} \approx 0.014$ even when all 8 are
equally good. The expression is the expected maximum of 8 draws from a
normal distribution, so it assumes approximate normality of the paired
differences and independence between candidates. Neither holds exactly here:
our resamples overlap, and candidates within a family are correlated. Both
departures make the true inflation smaller than this figure, so we use it as
an upper bound on how much a selected winner may owe to selection, not as an
estimate of it. That is the same order of magnitude as the effects being
chased, which is why a change that merely came first among several is held
to a stricter standard than one specified in advance.

\textbf{One thing the instrument still cannot see.} Every protocol above
compares systems built the same way. It does not protect against a
difference between how a system is measured and how it is finally used. In
our case a submitted prediction is the average of 5 fold models, while the
held-out prediction the cutoff was fitted on came from the single fold
model that had not trained on that post. \textit{No amount of resampling
would reveal a mistake that both sides of the comparison share.} The
calibration error described in the introduction, to which we return below,
reached the leaderboard through exactly that gap.

% =====================================================================
\section{Testing the Playbook}

This section carries the evidence the rule is drawn from. The 31
pre-specified comparisons drawn from the standard playbook fall into 5 mutually
exclusive verdicts, and \textbf{26 did not survive}. Of those 26: 12
produced no effect we could measure, 6 made a component measurably worse, 2
could not be resolved because the component they were measured on has no
floor, and 6 gave mixed or qualified outcomes, meaning they helped on 1
component and not another, or helped only under a condition we could not fix
in advance. The larger backbone is 1 of those 6: it helped once and then
stopped, which is why it is not counted among the survivors. All 31 are set
out one by one in Table~\ref{tab:playbook}. \textit{The 5 that survived
share a property that almost none of the 26 has.}

\begin{figure*}[!t]
\centering
\includegraphics[width=\textwidth]{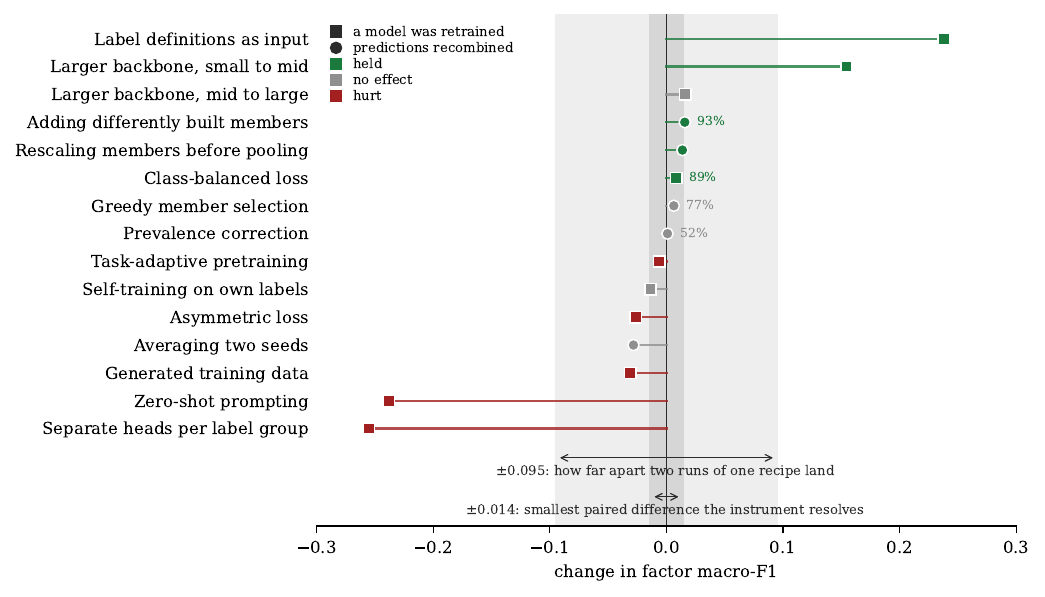}
\caption{Most of the standard playbook disappears into the noise floor.
The horizontal axis is the change in factor macro-F1 against the system
each technique was tested on, and each row is 1 comparison. Every
comparison we could place on the factor component is shown,
and color gives the verdict from
Table~\ref{tab:playbook}. Those verdicts follow
from the protocol of Section~\ref{sec:protocol}, in which a change is
accepted when it wins on a large majority of author-disjoint resamples,
rather than from magnitude alone; win rates are shown where we recorded them. The 2
shaded regions give scale. The outer is how far apart 2 runs of one
recipe land in absolute score, and the inner is the smallest paired
difference the pooled instrument can resolve. A paired comparison is more
sensitive than the outer band alone suggests, which is why several small
effects here are judged to have held while several larger ones are not.}
\label{fig:effects}
\end{figure*}

A word on how to read what follows. Every comparison changes one thing and
holds the rest fixed, and every offline number is read on author-disjoint
resamples
under the protocol of Section~\ref{sec:protocol}. Scores for the factor and
evidence components are not comparable with those for risk, so each result
names the component it belongs to. A verdict follows the win rate wherever
we have paired resamples, and the seed-range floor of the component
otherwise: 0.095 on factors, 0.0183 on risk, with any difference inside
that floor recorded as no effect whichever way the arithmetic fell. Neither
the evidence component nor the composite has a floor, so rows measured
there are marked unresolved, unless the effect dwarfs any floor we measured
anywhere or was read directly off the leaderboard. Two terms recur: a \emph{member} is one
trained model that contributes a score, and the \emph{pool} is the set of
members whose scores are combined into the factor output. The 7 families
are model capacity; more data, adaptation and input; the training objective
and sampling; ensembling and member combination; architecture and output
structure; decoding, rules and thresholds; and label definitions, prompting
and large language models. Table~\ref{tab:playbook} gives one row per
comparison, and Figure~\ref{fig:effects} plots every one we could place on
the factor component against that component's floor. The roughly 300 runs
behind the table include the seed replications and sweeps each comparison
required.

\begin{table*}[!t]
\centering
\caption{26 of these 31 techniques did not survive, and all 5 that survived consume something this task specifically supplies. Each row gives what the
literature reports for a technique and what we measured here. Verdicts
follow the rules set out in the text: a win rate where we have paired
resamples, the seed-range floor of the component otherwise, and unresolved
where the component has no floor. Scores are not comparable across
components, so every row names the one it was measured on. Scores are that component's own metric unless the row says composite.}
\label{tab:playbook}
\footnotesize
\setlength{\tabcolsep}{4pt}
\renewcommand{\arraystretch}{1.05}
\begin{tabular}{@{}p{0.215\textwidth}p{0.225\textwidth}p{0.300\textwidth}p{0.185\textwidth}@{}}
\toprule
Technique & Reported elsewhere & Measured here & Verdict \\
\midrule
Task-adaptive pretraining & $+4.4$ F1 at this data size \cite{gururangan2020dapt} & $-0.0062$ as a pool member (factors) & no effect \\
Larger backbone (factors) & reliable \cite{he2019bagoftricks} & $0.4495 \to 0.6038 \to 0.6199$ & helped once, then stopped \\
Larger backbone (risk) & reliable \cite{he2019bagoftricks} & $0.8001 \to 0.7902$ & no effect \\
Generated training data & worse on subjective tasks \cite{longpre2020augmentation} & $0.5900 \to 0.5590$ (factors) & no effect \\
Self-training on own labels & up to $+2.6$ accuracy reported \cite{du2021selftraining} & $0.6038 \to 0.5901$ (factors) & no effect \\
Retrieval of similar posts & untested for this setting & lost on all 24 categories (factors) & hurt \\
Longer input window & contested & no variant selected (evidence) & no effect \\
Class-balanced loss & small gains \cite{cui2019classbalanced} & $+0.0084$ macro (factors), 89\% of resamples & held \\
Focal loss & contested \cite{lin2017focal} & 1 seed $+0.0133$, 2 seeds $-0.0061$ and $-0.0340$ (risk) & did not replicate \\
Asymmetric loss & reported for long-tail & $-0.0263$ macro (factors) & no effect \\
Difficulty-weighted sampling & widely used for imbalance & $0.7576 \to 0.7450$ (risk) & no effect \\
Architecturally diverse ensemble & mechanism contested & $+0.0157$ macro (factors), 93\% of resamples & held \\
Seed ensembling & most reliable in the canon \cite{lakshminarayanan2017ensembles} & $0.6038, 0.5782 \to 0.5755$ (factors) & no effect \\
Greedy ensemble selection & established in AutoML & $+0.0063$ vs $+0.0157$ pre-specified (factors) & lost to a fixed set \\
Leave-one-out member pruning & established & $+0.0010$ evidence, $-0.0072$ risk & no effect on risk, evidence unresolved \\
Hard majority voting & widely used default & $k{=}3$ $0.7953$ vs $k{=}2$ $0.8045$ (evidence; $k$ = taggers that must agree) & no band, unresolved \\
Stacking on model outputs & established in competitions & $0.7967 \to 0.7822$ (risk), 96 fitted parameters on 1,635 posts & no effect \\
Shared encoder, separate heads & efficiency and transfer & $0.4220$ against a $0.6770$ pool (factors) & hurt \\
Concept bottleneck & interpretability at no cost & $0.7358$; $0.7403$ with concepts zeroed (risk) & no measurable contribution \\
Ordinal regression head & rank consistency \cite{shi2023corn} & $0.7579$ flat vs $0.7363$ ordinal (risk) & hurt on one architecture \\
Post-hoc symbolic rules & common in clinical NLP & $0.7858 \to 0.7841$ out of fold (risk) & no effect \\
Rules scoped to the evidence & not previously reported & $+0.0018$ risk weighted F1, 99\% of resamples & held \\
Per-category thresholds & established theory \cite{lipton2014thresholding} & inflates in-sample score by $0.0684$ (factors) & in-sample bias \\
Prevalence correction & established, zero parameters & $+0.0009$ macro (factors), 52\% of resamples & no effect \\
Threshold estimator swap & plausible refinement & $-0.0261$ and $-0.0287$ composite, on the leaderboard & hurt \\
Label definitions as input & $+10$ to $+18$ vs.\ classical models \cite{laurer2024nli} & $0.3795 \to 0.6172$ (factors) & held, largest in the playbook \\
Same, applied to evidence & not previously reported & $0.4830$ vs $0.7787$ (evidence) & hurt \\
Zero-shot prompting & underperforms fine-tuning \cite{yin2019entailment} & $0.4040$ vs $0.6420$ (factors) & hurt \\
Parameter-efficient tuning & near-neutral vs full tuning & $0.7694$ vs $0.7646$ (risk) & no effect \\
Routing one class to a separate model & not previously reported & $0.7536 \to 0.7588$ composite & held \\
LLM as reranker & contested & no configuration beat the tagger vote (evidence) & no band, unresolved \\
\bottomrule
\end{tabular}
\end{table*}

\textbf{Adding capacity helped once, and then stopped paying.} Moving from
the smallest encoder to the mid-sized one \cite{he2023debertav3} raised the factor score from
0.4495 to 0.6038, by a wide margin the largest effect any change in model
size produced. The next step up, to a model roughly twice as large,
returned 0.6199. That second gain of 0.0161 fell inside the factor seed range and
we do not count it. The same ladder run on the risk component gave
0.8001 for the middle size and 0.7902 for the largest, a difference of
0.0099 against a risk seed range of 0.0183. The second step up in capacity produced
nothing we can measure on either output, and neither of the largest models
reached the submitted system. The honest
summary is not that capacity never helps, but that here it helped once,
steeply, and then the returns fell below what this corpus can resolve.

\textbf{Synthetic and pseudo-labeled supervision did not help, and generated
data taught a shortcut instead.} We tried 3 ways of manufacturing more
supervision, none of them additional human labels, and none of them paid. Labeling additional posts with a large
language model and adding them to training moved the factor score from 0.5900 to 0.5590, a difference inside the factor
seed range and therefore no effect we can resolve. Using our own
validated model as the teacher instead, which removes the mismatch between
an external annotator's judgments and the codebook's definitions, gave
0.5901 against an incumbent of 0.6038, the incumbent being the best system
we had at the time of that comparison. That difference sits inside the factor
seed range, so we record no effect. Text generated to fill out the rarest
categories carried surface regularities the real corpus never showed:
openings of one particular form appeared in 16.8 percent of generated
posts and in none of the real ones. That observation matters beyond the
score, because a model able to separate real from generated text on a
surface cue alone is not learning the distinction we intended to teach it.

\textbf{Only 1 of the 4 changes to the training objective held.} Focal
loss, which concentrates training on the examples a model finds hardest,
was swept over 6 settings on the risk component. Only one setting beat
plain cross-entropy, at 0.7904 against 0.7771. Re-running that single
setting with 2 further seeds gave 0.7710 and 0.7431, and the setting the
original paper recommends scored below the plain loss at every seed we
tried. A single-seed result that does not survive 2 replications is the
clearest illustration in this paper of why the seed range of
Section~\ref{sec:protocol} had to be established first. An asymmetric loss built for long-tailed problems moved the factor score by
$-0.0263$ macro, again inside the seed range. Difficulty-weighted
sampling, which draws hard examples more often, scored 0.7450 against
0.7576 for uniform sampling, a difference of 0.0126 that sits inside the
risk seed range, so we record it as nothing rather than as a loss. A
class-balanced loss, which weights each category by how rare it is, gained
$+0.0084$ macro on the factor component, won on 89 percent of resamples,
and beat its own matched control on all 5 folds.

\textbf{Ensembling paid only when the members differed, and that is a
measured property of this system rather than a rule about ensembles.} Averaging 2
runs of one recipe produced no measurable gain. The 2 runs scored 0.6038
and 0.5782, and their average scored 0.5755. The average fell below both
members, but the gap sits well inside the factor seed range, so we record only
that the most reflexive move in the playbook bought nothing here. What did
pay was difference rather than quality. A pool of 9 members sharing one
architecture had stopped improving, and no further member of that same
architecture, added or removed, changed its score. Two members built on different architectures
reopened it, together worth $+0.0157$ macro, winning on 93 percent of
resamples. Those 2 make errors that correlate less with the pool than
any existing member does, one of them because it is trained jointly on all
3 outputs rather than on factors alone, and it is that decorrelation
rather than their own accuracy that pays. The weaker of the 2 by
standalone score still earned a place. A selection rule based on standalone quality would
have dropped it. Letting a greedy
search choose the members instead of fixing them in advance did worse,
gaining 0.0063 on 77 percent of resamples, which is the signature of a
search fitting its own sample.

\textbf{The failures are not an artifact of a strengthening
baseline.} Each comparison was run against the incumbent of its day, and
that incumbent improved over the campaign, so a technique tested late
faced a harder baseline than one tested early. One version of that concern can be
ruled out: that a technique failed only because the particular incumbent it
happened to meet had already absorbed the gain it offered. If that were so,
the same technique should still add something to a different and stronger
system assembled from other parts. Taking 5 techniques
that failed at various points, including all 3 variants of generated
training data and task-adaptive pretraining,
we added each to the final 11-member pool and measured each one's leave-one-out contribution to the pool. All 5
contributions were negative: $-0.0040$,
$-0.0044$, $-0.0019$, $-0.0050$ and $-0.0062$. Every one is listed by the
procedure as a member whose removal improves the pool. \textit{A technique
that failed early fails against the finished system too.}

\textbf{The one trick with published support at our data size still failed.}
Task-adaptive pretraining is the single item of the playbook whose original
paper reports a gain on a corpus the size of ours, 4.4 F1 on 1,688 examples
against our 1,635 \cite{gururangan2020dapt}. Corpus size is the only dimension
on which the 2 settings match; the task, domain, objective and metric all
differ. \textit{If any
borrowed number should have carried over, it is that one, and it did
not.} We continued pretraining the backbone on the
task's own text and trained a factor member on the adapted model. It scores
0.5950 alone against 0.6132 for the plainest member in the pool, and adding
it to the pool costs 0.0062.

\textbf{Adding structure to the model did not help.} All 3 attempts to give the
model more of the task's structure failed to raise the score. Splitting
the output head so that risk factors and protective factors were predicted
by separate sub-networks scored 0.4220 against the 0.6770 of the pool it
was meant to improve. A concept bottleneck, which forces predictions
through an intermediate layer of named clinical concepts, scored 0.7358.
The same architecture with every concept set to zero scored 0.7403. The
head keeps a direct path from the encoder alongside the concept layer, so
zeroing the concepts leaves that path intact, and the result says the
concept channel contributed nothing measurable on top of it. An ordinal head, often
treated as mandatory for an ordered label like this one, scored 0.7363
against 0.7579 for a flat classifier over the same 4 levels, although the
same head helped on a different architecture.

\textbf{Label co-occurrence had little to give under this test.} One family of
techniques tries to exploit the fact that categories co-occur, predicting each
label partly from the others. We measured the most that this could possibly buy. Given the true values of the other 23 categories and no
access to the post at all, a classifier predicting the remaining one
reaches 0.2732 macro-F1, against 0.7045 for the deployed system that reads
the text. The categories do co-occur, at 2.92 labels per post, but not
informatively enough for one to be inferred from the rest. Any technique
whose gain must come through that channel is bounded by this number before
it is written.

\textbf{Rules tuned on the public split did nothing on posts they had not
seen.} A layer of hand-written symbolic rules applied after the model, of
the kind common in clinical text processing, scored 0.7841 against 0.7858
for the model without the rules when measured out of fold. The rules had
been tuned on the public evaluation split, where they gained a visible
amount; on the 1,635 posts scored out of fold, each by the 1 fold model that had
not trained on it, the difference was 0.0017, a tenth of
the risk seed range and indistinguishable from a change of seed. The scoped
rules of Section~\ref{sec:calibration} move the score by a similar amount,
and what separates them is the instrument rather than the magnitude: the
scoped version won on 99 percent of 50 author-disjoint resamples, and this
one was never put on that instrument at all. \textit{This is the clearest case in the paper of a
technique that works where it was fitted and nowhere else}, and it is the
setup for the one rule family that did survive.

\textbf{The decoding choices mattered as much as the modeling ones.}
Turning 24 probabilities into 24 decisions requires a cutoff for each.
Fitting those cutoffs on the same posts used to score them inflates the
result by 0.0684, larger than all but 2 of the modeling effects we measured
on that component. What fails is the fitting, not the idea: the shipped
system does use a cutoff per category, fitted out of fold. Correcting the predicted rate of each category toward its rate
in the training data, which fits nothing at all, appeared to be worth
$+0.0235$ macro on 100 percent of resamples when the rates were taken from
the whole corpus. Fitting those rates on one set of authors and scoring on
another, which is what deployment actually does, reduces it to $+0.0009$
on 52 percent of resamples. The first number is an artifact of letting the
correction see the posts it is scored on, and the second is the one we
report.
Two more principled estimators of the same cutoffs, both of which read
better than the incumbent on our offline instrument, lost 0.0261 and
0.0287 when submitted. Those 2 losses are what prompted the protocol of
Section~\ref{sec:protocol}, and they are why this paper treats the
cutoff-fitting procedure as an experimental variable in its own
right rather than as a detail of implementation.

\textbf{Telling the model what a label means produced the largest gain in
the playbook.} Replacing 24 anonymous output slots with 24 written
clinical definitions, and asking the model whether each definition
describes the post, moved the factor score from 0.3795 to 0.6172 with the
underlying model, its size, and its training data all unchanged. No change
in model size came close. As described in the introduction, the same move applied to the evidence output reached 0.4830 against 0.7787 for an
ordinary tagger, because that part of the task has no definition to
supply. The technique is not ours and neither is the explanation for the
size of the effect: it is reported to help most in exactly this data
regime, and this corpus falls inside it.

\textbf{Reading more of the post bought nothing.} Only 2.0 percent of
marked evidence falls beyond the 512-token window an ordinary encoder
reads, so the headroom a longer window could recover is small by
construction. We built variants reading 256, 384 and 1{,}024 tokens, both
shorter and longer than the default, and offered all of them to the
member-selection procedure, which chose none.

\textbf{In the configuration we tested, prompting a large model lost to
fine-tuning an encoder.} A 72-billion-parameter instruction-tuned model
\cite{qwen25} prompted without training scored 0.4040 on the factor
component against 0.6420 for the definition-conditioned encoder. The
comparison varies model family, optimization regime and definition
conditioning together, so it bounds this configuration rather than isolating
any 1 of the 3. Fine-tuning a
14-billion-parameter model on the same 24 outputs, without the definitions,
does not settle the question, because the answer depends on the training
budget: at 3 epochs it trails the definition-conditioned encoder by 0.0106
and at 6 it leads by 0.0399, with the folds still improving when we
stopped. We therefore report it in the limitations rather than as one of
the 31.

\textbf{The 26 failures share nothing except what they did not ask of
this task.} The 26 negative rows have no single cause in common. They
fail on capacity, on data, on the training objective, on ensembling, on
architecture, on decoding and on prompting alike, which is why no narrower claim about one
family of techniques would cover them. \textit{What almost all of them
share is that they asked nothing of this task in particular.} That is a
statement about a regime rather than about a corpus, and it is one no
previous audit was positioned to make, because every earlier one was run
where the data is plentiful, the output is single, and the metric follows
the frequencies rather than protecting the rare category.

% =====================================================================
\section{Task-Conditioned Technique Selection, and the Fix That Was Not a Technique}
\label{sec:calibration}

This section sets out the 2 methodological findings the audit produced. The
first is a rule for deciding which borrowed techniques are worth trying at all.
The second is a defect in how we turned scores into decisions, which cost us
more than any modeling decision we made.

\textbf{Task-conditioned technique selection.} \textit{We did not begin with this
principle; it is what the results left behind.} We expected the playbook to
hold, and set out to confirm it. What the 31 comparisons left instead was a single regularity: \textit{every
technique needs some particular piece of task knowledge, and a technique
paid here exactly when this task supplied it.} Written definitions consume a codebook. A class-balanced loss
consumes real imbalance. A diverse ensemble consumes errors that differ between
members. Rules scoped to a clause consume an annotated clause. Routing consumes
an identifiable weak class. Each of those 5 demands is a piece of task knowledge in the sense of
Section~\ref{sec:task}. This task supplies all 5, and all 5 of those
techniques held. The 26 that did not hold asked for something
this task does not have, or asked for nothing in particular.

As a procedure it is 4 steps, of which the first 3 come before any GPU time.
Name the knowledge or structure the technique needs. Ask whether the task
supplies it. Then measure whether what it supplies is informative, because a
task can contain something and still not contain enough of it to work with:
the 24 factors do co-occur, but an oracle given the true values of the other
23 reaches only 0.2732, and only 2.0 percent of marked evidence falls beyond
a 512-token window. Only then run the controlled comparison, and keep the
component only if that comparison supports it. Table~\ref{tab:resource} applies those 4 steps to the audit in retrospect,
and carries them one step further: each row runs from the knowledge or
measured property, through the modeling decision it licenses and the
measurement that tested it, to whatever survived into the shipped system.
Nine of the 15 chains terminate in nothing. That reduction is what the audit
bought, and it is why we describe the result as removing moving parts rather
than as ranking techniques. We give the rule a name to make it reusable, not to claim the
observation is ours. It is a hypothesis for the next corpus of this shape, not
a law. The paired experiments below are its first evidence.

\begin{table*}[!t]
\centering
\caption{The audit as a chain. Each row runs from what the task supplies,
through the modeling decision that knowledge licenses and the measurement that
tested it, to what remains in the shipped system. Six chains terminate in a
component and 9 terminate in nothing: that is the sense in which the audit
removed moving parts rather than ranked techniques. Five of the 6 are the
surviving playbook techniques; the sixth ($\dagger$) is the calibration repair,
which is not a playbook technique at all but the measured failure mode of
Section~\ref{sec:calibration}. Evidence figures are those of
Table~\ref{tab:playbook} and name the component they were measured on.}
\label{tab:resource}
\footnotesize
\setlength{\tabcolsep}{4pt}
\renewcommand{\arraystretch}{1.15}
\begin{tabular}{@{}p{0.225\textwidth}p{0.200\textwidth}p{0.275\textwidth}p{0.190\textwidth}@{}}
\toprule
Task knowledge or measured property & Modeling decision it licenses &
Experimental evidence & In the shipped system \\
\midrule
\multicolumn{4}{@{}l}{\textit{Prior knowledge: on the page before any model is trained}} \\
\addlinespace[2pt]
A codebook defining all 24 factors in prose & Factor members as entailment:
the definition is the hypothesis & $0.3795 \to 0.6172$ factors, the largest
effect in the audit & All 11 factor members \\
Label counts from 8 to 745 positives & Class-balanced loss on one member &
$+0.0084$ macro (factors), 89\% of resamples & 1 of the 11 members \\
An annotator-marked clause behind every risk judgment & Symbolic rules
restricted to that clause & $+0.0018$ risk weighted F1, 99\% of resamples;
rule precision $0.284 \to 0.667$ & 3 rules, firing on 8 posts \\
\addlinespace[3pt]
\multicolumn{4}{@{}l}{\textit{Empirical knowledge: only by measurement}} \\
\addlinespace[2pt]
Errors decorrelate across encoder families & Add 2 members built on other
architectures & $+0.0157$ macro (factors), 93\% of resamples & 2 of the 11
members \\
\emph{attempt} is the class the chain is weakest on & Route that one class to
a reasoning model & $0.7536 \to 0.7588$ composite & The 72B routing step \\
Fold-averaging narrows each member's range, unevenly (0.804--0.970)
$\dagger$ & Rescale each member onto its held-out range before pooling &
$+0.0137$ factors, $+0.0041$ composite & The rescaling step before the pool \\
\addlinespace[3pt]
\multicolumn{4}{@{}l}{\textit{Asked for something this task does not supply}} \\
\addlinespace[2pt]
The codebook defines no spans & The same definition conditioning, on evidence
& 0.4830 against 0.7787 (evidence) & removed \\
Errors of a 9-member family already correlate & Further members of that same
family & 0.6038, 0.5782 $\to$ 0.5755 (factors) & removed \\
No information the corpus does not already hold & Generated data;
self-training on our own labels & $0.5900 \to 0.5590$; $0.6038 \to 0.5901$
(factors) & removed \\
No unlabeled in-domain text at scale & Task-adaptive pretraining & $-0.0062$
as a pool member (factors) & removed \\
2.0\% of marked evidence beyond 512 tokens & Longer input window & no variant
selected (evidence) & removed \\
No clinical concept the encoder lacks & Concept bottleneck & 0.7358; 0.7403
with concepts zeroed (risk) & removed \\
Too few rows to fit 96 parameters & Stacking on model outputs & $0.7967 \to
0.7822$ (risk) & removed \\
Too little usable ordinal structure & Ordinal regression head & 0.7579 flat
against 0.7363 ordinal (risk) & removed \\
The clause exists, but the rule never reads it & Symbolic rules over the whole
post & $0.7858 \to 0.7841$ out of fold (risk) & removed \\
\bottomrule
\end{tabular}
\end{table*}

\textbf{The same technique, with and without the knowledge.} So far the rule
only explains the results it was drawn from. That is interpretation, not
evidence, and an explanation written after the fact is worth little. But on 4
occasions we did better. We ran the same intervention twice, changing only
whether this task supplied the knowledge that intervention needs, and the effect
changed with it.

Definition conditioning moved the factor score from 0.3795 to 0.6172, the
largest effect in the audit. Applied to the evidence output it scored 0.4830
against 0.7787 for an ordinary tagger. The intervention is the same; the
codebook defines the 24 factors and defines no spans.

Ensemble diversity across architectures gained 0.0157 macro on 93 percent of
resamples, while adding further members of the same family to the same pool
moved 0.6038 and 0.5782 to 0.5755. The technique is the same; only the members'
errors differ. The 2 members that did reopen the pool were built differently,
and one is roughly a quarter the size of those it joined and scores 0.4078 alone
against their 0.61 to 0.64. Its contribution is not the largest, so we do not
claim that weakness itself is an asset; what the pool needed was a member whose
errors fall elsewhere.

Symbolic rules applied to the whole post moved 0.7858 to 0.7841, nothing; the
same rules scoped to the clause the evidence taggers marked gained 0.0018 risk
weighted F1 on 99 percent of resamples, raised the precision of the
completed-act rule from 0.284 to 0.667 on the posts where its trigger phrase
appears, and stopped needing a fitted margin at all. The rules are the same;
only the text they read changed. That scoping fixes 7 posts and breaks 1, and
applying it to the 2 other candidate rules we had written made both worse,
costing 0.068 and 0.066 F1, because those 2 look for context the clause
deliberately excludes.

Routing completes the set: sending the attempt class to a separate model gained
on the class the main model is weakest on, and the same routing turned negative
when pointed at a class that was not weak. \textit{Four techniques were each run
with that knowledge present and absent, and all 4 either changed sign or lost
their effect: we have a negative control for 4 of the 5.} \textbf{One of the 5
survivors has no such control, and we do not claim one}: the class-balanced loss
would need a comparable task without a long tail, and we have none.

\textbf{What is new here.} We do not propose another model component. We
propose a way of deciding which existing components a task justifies. Name the task knowledge a technique operates on. Establish that the task
supplies enough of it to matter. Only then spend the compute. The 4 matched comparisons above are controlled evidence
for that procedure. The 31-technique audit measures how often the conventional
playbook skips it. \textit{The contribution is the selection
procedure and the evidence for it, not the components it happened to select.}

\textbf{Deployment-consistent calibration: the largest gain came from
correcting a mismatch between how we measured and what we shipped.} The
mismatch is a cross-validation-to-ensemble shift in the score distribution:
thresholds chosen on the predictions of a single fold model were applied to
an average of 5, which is a narrower distribution, and narrower by a
different amount for each member. Nothing about the models was wrong. The cutoffs
were fit on one set of numbers and applied to another. Each of the 11
members predicting the 24 factors returns a score per category, and those
scores are combined before a cutoff turns them into decisions. Each member is itself an average of 5 models, one
trained on each fold of the data. The cutoffs, however, were chosen on
held-out predictions, where a post is scored by the single fold model that
did not train on it. Averaging pulls scores toward the middle. So the
spread of the numbers reaching the cutoff at test time is narrower than the
spread the cutoff was chosen against, and the amount of narrowing differs
from member to member.

\begin{figure}[t]
\centering
\includegraphics[width=\columnwidth]{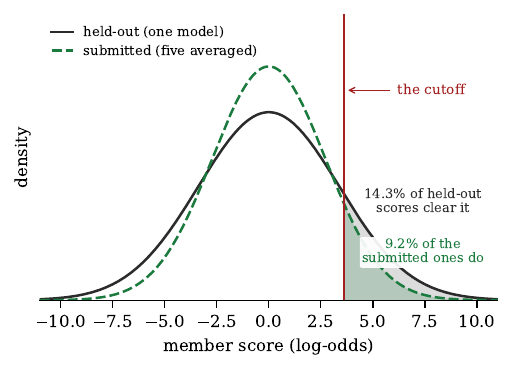}
\caption{Why the cutoff ended up in the wrong place. The 2 curves show the score distribution of one
ensemble member, drawn with its measured
spreads: 3.374 on held-out predictions, where each post is scored by the
single fold model that did not train on it, and 2.714 on the submitted
predictions, which average 5 fold models. The cutoff was chosen against
the wider distribution and applied to the narrower one, so the share of
scores clearing it falls from 14.3 to 9.2 percent without anything about
the member having changed. The amount of narrowing differs from member to
member, which is what shifts the balance of the pool.}
\label{fig:calibration}
\end{figure}

That narrowing does more than move the cutoff. The members are combined by
an unweighted mean of their log-odds, on the scale of $\log(p/(1-p))$, and
an unweighted mean still gives each member influence in proportion to how
widely its own scores range. A member whose scores barely move contributes
almost nothing to the variation of the sum. So a member whose scores
narrowed a lot lost influence, and a member whose scores narrowed little
gained it, without any weight being changed. The cutoffs were fit when the
pool had one composition and applied when it had another. The narrowing is
uneven as well as real: across the 11 members the ratio of test spread to
held-out spread runs from 0.804 to 0.970, so one member lost a fifth of its
range while another lost 3 percent.
Figure~\ref{fig:calibration} shows the shift for a single member.

The correction is to map each member's test scores back onto the range its
own held-out scores occupied, and only then combine the members. It fits
no parameters, reads no labels, and requires no retraining. It was worth
$+0.0137$ on the factor score and $+0.0041$ on the composite. As an increment
to a system that was already finished, \textbf{it was worth more than any
modeling decision we made}. The comparison that matters here is against the
same finished pool: architectural diversity is the larger number at $+0.0157$,
but it was measured when 2 members joined a pool of 9, not added to the
11-member pool the system shipped with. The only other number larger is the
0.2377 from definition conditioning, but it is not the same kind of quantity: it was measured
against a naive baseline rather than added to a tuned system, and moving a
system from bad to good is easier than moving it from good to better.

\textbf{Why our offline checking could not have found it.} The defect
lives in the difference between the held-out predictions and the submitted
ones. Any offline comparison scores both the old system and the new one on
the held-out predictions, so both sides of the comparison inherit the
defect equally and it cancels. This is a general hazard rather than a
local one. It will arise wherever cutoffs, weights, or thresholds are fit
on cross-validated predictions and applied to a model trained on all the
data, and wherever members are combined by a rule sensitive to the range
of their outputs. \textit{That covers a large share of shared-task submissions and of
production retraining.} The variance-reduction mechanism behind it is known
\cite{collell2018threshold}. We are not aware of prior work that
characterises it as a deployment failure which offline validation cannot
see, though we claim only that our own search did not find one, and that is
why we state it at length here rather than leaving it to an implementation
note.

\textbf{What the surviving techniques have in common.} Each of the 5 survivors exploits a specific piece of task knowledge, and the
5 divide into the 2 kinds named in Section~\ref{sec:task}. Three are \emph{prior} knowledge, available before any model is
trained: the written definitions in the codebook, the steeply unequal label
distribution, and the annotator-marked justifying clause. Two are
\emph{empirical} knowledge, discovered only by measurement: that errors differ
across model families, and that one risk class is where the chain is weakest.
The distinction matters because only the first kind can be read off the task
before spending compute. Written
definitions worked because the categories come from a codebook. A
class-balanced loss worked because the label distribution really is
steeply unbalanced. Members built on different architectures worked
because their errors correlate less with the pool than any same-family
member's do. Scoping symbolic rules to the evidence clause worked because
the annotators marked that clause for exactly the reason the rule looks
for. Routing one risk class to a separate model
worked because that class is the one the main model is weakest on. None of the 5 is a method of ours. Three come straight out of the same
literature that supplied the 26 that failed, and the other 2 are ordinary
moves scoped to this task. \textit{What separates the survivors from the
failures is not where they came from, but whether this task contains the
thing each one is built to consume.}

% =====================================================================
\section{The System, and What It Does Not Contain}
\label{sec:system}

\begin{figure*}[!t]
\centering
\includegraphics[width=\textwidth]{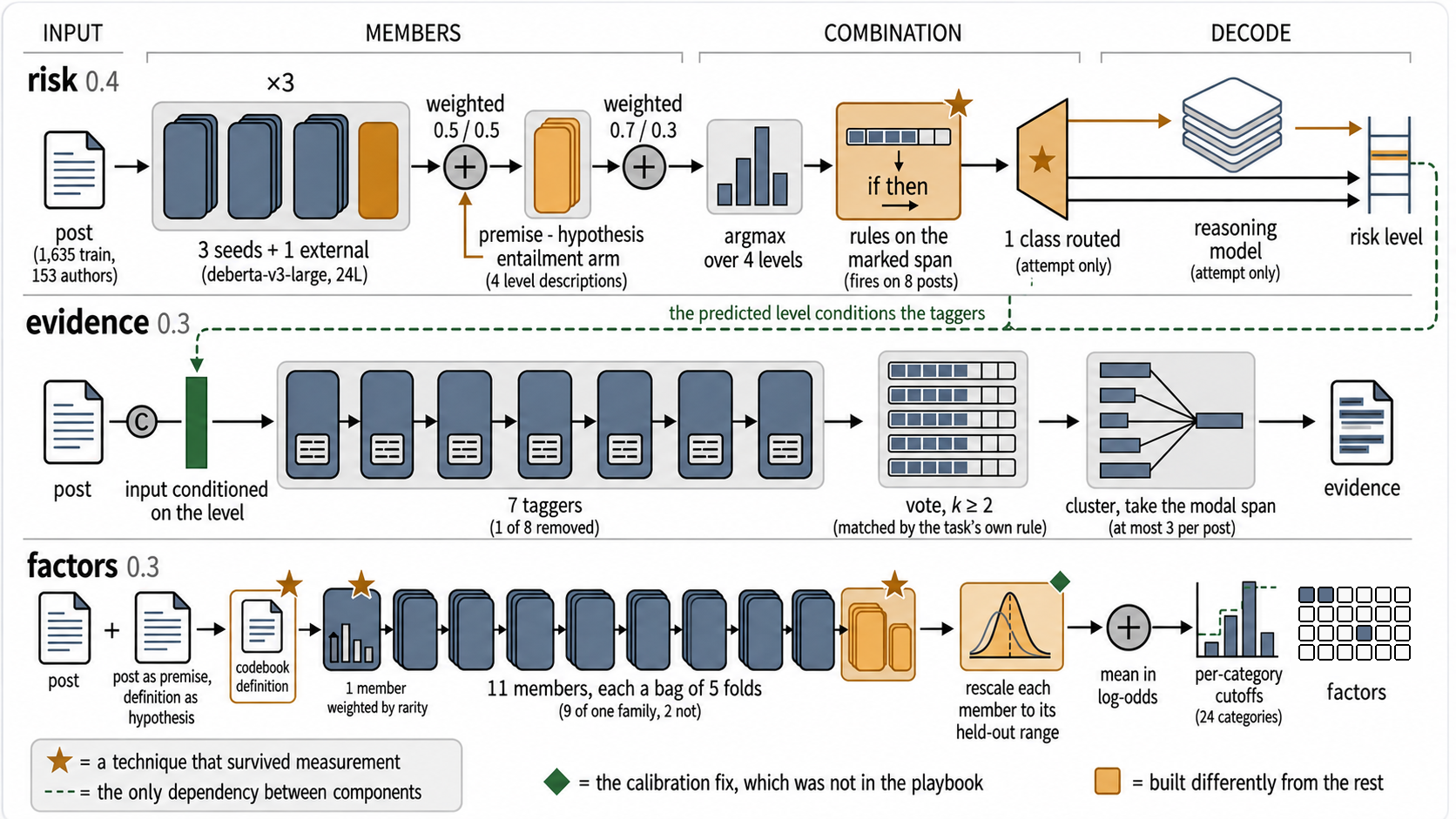}
\caption{The submitted system. The post is read independently by 3
components. Three dependencies run between them, in a fixed order. The blended risk
level conditions the evidence taggers, the rules then read the clause those
taggers marked and may raise the level, and a final level of indicator empties the evidence output entirely. That
last rule is worth stating precisely: it is right for 588 of the 611
lowest-level posts in the annotated corpus and wrong for 23, and we keep it because
an empty prediction scores 1.0 against empty gold, so the rule buys far more
on the 588 than it forfeits on the 23. The factor pool does not see the risk level, for
a reason given in the text. The 5 starred blocks are the 5 techniques that survived
Table~\ref{tab:playbook}, and the separately marked rescaling step is the
calibration fix; Table~\ref{tab:resource} traces all 6 from the task
knowledge that licensed them through to the block drawn here. The remaining
departures from plain fine-tuning are set out in Table~\ref{tab:system}.}
\label{fig:system}
\end{figure*}

Figure~\ref{fig:system} shows the submitted system, which we call TRIDENT
after the 3 outputs it has to produce. The name is a label rather than a
claim: what is worth attention is not the arrangement, which is ordinary, but
how each part of it was chosen. We set it out in full here rather than compressing it into an implementation
note, because the final architecture is what the audit produced: unsupported
modeling choices pruned away, and only those departures from plain fine-tuning
retained that a piece of task knowledge or a measured failure mode justifies. Almost every non-obvious choice in it is the surviving end of a comparison
in Table~\ref{tab:playbook}, and everything else about it is plain because
nothing in the task justified making it otherwise. The 5 starred blocks in Fig.~\ref{fig:system} are the 5 survivors and the
rescaling step beside them is the calibration fix; Table~\ref{tab:system}
holds the 4 further departures our rule does not account for, and everything
else is plain, which is the record of what did not survive. \textit{What we would put forward is not an
extra layer of complexity, but a procedure for deciding which complexity this
task justifies.} TRIDENT is what that procedure returned, not something designed in advance.

\textbf{Simplification through task knowledge is what the system is, not a
side effect of building it.} We implemented and measured 31 techniques and shipped few of them. The
finished system has no auxiliary head, no ordinal output, no concept
bottleneck, no generated training data, no self-training, no retrieval, and
no task-adaptive pretraining of our own. Each of those absences is a component we built, scored and took out again, and each is a row of Table~\ref{tab:playbook}. What is left departs from plain fine-tuning in only 10 places, and Table~\ref{tab:resource} traces 6 of them back to a specific piece of task knowledge. \textit{The contribution is not a more elaborate architecture. It is that reading the task first told us which components a system of this kind actually needs.}

\textbf{The risk chain.} The risk output is produced by a chain of 3
stages followed by 2 adjustments. The first stage is an ensemble of 3 copies of a mid-sized encoder,
fine-tuned on the 4 levels and differing only in random seed, whose
predicted distributions are averaged. We kept it although seed ensembling showed no measurable gain on the factor
component, and this is the 1 departure the audit does not justify. Averaging
3 seeds triples the inference cost of this stage. We accepted that because
it reduces the variance of a single training run, which is a reason of a
different kind from the ones this paper argues for, and we flag it rather
than dress it up. The second brings in a model built independently, on a publicly released
encoder already pretrained on mental-health text (not the task-adaptive
pretraining we ran ourselves and rejected), and with 3 output heads rather
than 1, and averages it with the first at equal weight; it is the risk component's counterpart to the architectural
diversity that mattered so much on the factors. The third blends the
result, at weights 0.7 and 0.3, with a member that treats the task as
entailment, scoring each level's written description against the post
rather than learning the 4 levels as opaque classes. The blended
distribution is read off at its maximum, and the adjustments follow. A
small set of symbolic rules, each of which reads only the clause the evidence
stage has marked rather than the whole post, raises the level where a
completed act is described. Then posts that a separate reasoning model
judges to describe an actual attempt, and which the chain has not already
placed at the lowest level, are moved to the highest. That reasoning model is
Qwen2.5-72B-Instruct, used as released and prompted rather than fine-tuned,
loaded in 4 bits and shown one post at a time together with the 4 level
definitions; it sees no training labels and no other post. That last step is
the routing result of Table~\ref{tab:playbook}: it is applied to a single class, because the same move applied to a different class turned
negative. The chain scores 0.8203 weighted F1 on the blind leaderboard set and is
the strongest of the 3 components.

\textbf{The evidence panel.} Evidence is a span-extraction problem with an
unusual scoring rule, since a predicted span counts as correct if it
contains the annotator's span or is contained by it, and in either case is
no more than 3 times its length. That rule tolerates imprecise boundaries and punishes
choosing the wrong sentence, so the panel is built to agree on which
clause matters rather than to agree on where it ends. Seven taggers each
propose spans. They share an architecture, but each receives the risk
level the chain predicted as part of its input, so a tagger reading a post
placed at the highest level looks for different words than one reading the
same post placed lower. A candidate survives when at least 2 of the 7
propose something that the task's own matching rule would count as the
same span, which makes the vote consistent with the metric instead of with
string equality. Surviving spans are clustered by that same rule, each
cluster is emitted in its most frequently proposed form, and at most 3
spans are returned per post. An eighth tagger was trained and is not in the
panel: leave-one-out testing found the panel improved by 0.0010 when it
was removed, though with no null band for the evidence component we
cannot call that result resolved.

\textbf{The factor pool.} The 24 factors are predicted by 11 members, each
of which is itself an average over 5 models trained on different folds, so
the factor output alone is 55 forward passes per post. The system is plain
in the sense that almost nothing in it departs from ordinary fine-tuning,
not in the sense that it is cheap to run; what the audit removed was
unjustified structure, not compute.
Every member is formulated as entailment rather than classification: the
post is the premise and the codebook's own written definition of the
category is the hypothesis, so the model is told what hopelessness means
instead of being left to infer it from the examples alone. This is the largest capability gain in the
paper, though not the biggest single improvement to the finished system. Of the 11 members, 9 share one encoder family
and 2 deliberately do not, one of them roughly a quarter the size of the
rest and weaker than all of them alone; those 2 are the only additions
that ever moved a pool that 9 same-family members had closed. One member
is trained with a loss that reweights each category by how rare it is.
Before the members are combined, each one's test scores are mapped back
onto the range its own held-out scores occupied, which is the correction
of Section~\ref{sec:calibration} and the biggest single improvement we made.
The rescaled members are averaged in log-odds, on the scale of
$\log(p/(1-p))$ rather than in probability. Each category is then cut
at its own threshold, fitted out of fold. The pool scores 0.7045 macro-F1 on the 378-post blind leaderboard set.

\textbf{The dependencies run in a fixed order, and we tested the reverse.} The blended risk level conditions the evidence taggers, the
span-scoped rules then read the clause those taggers marked and may raise
the level, and when the final level is the lowest the evidence output is
emptied outright.
That second rule looks aggressive and is not: as Table~\ref{tab:data}
shows, the annotators left the evidence field empty in 588 of the 611
lowest-level posts and in none of the 1,024 posts at any other level, so
emptiness and the lowest level are very nearly the same event in the gold
data. We tested letting the evidence output
decide the level outright and rejected that; the rules use the marked
clause only to restrict where they may fire, not to overrule the chain. We also tested the more tempting idea that the risk
evidence should inform the factors, on the reasoning that the annotators
had already marked the most diagnostic words in the post. That idea is false.
Training a factor model on the gold risk spans alone reaches 0.1897, and
training it on random text of the same length from the same posts reaches
0.1900. \textit{The words that justify a risk level carry no more factor signal
than an arbitrary passage of equal size}, which is why the factor pool in
Fig.~\ref{fig:system} reads the post directly and ignores everything the
risk chain produces. The 3 outputs are scored together, but the factor pool and the risk chain are nearly independent problems that happen to share an
input.

\textbf{Every departure from the plain recipe traces to a measurement.}
Table~\ref{tab:resource} already carried the 6 departures a piece of task
knowledge licenses, from that knowledge through to the component.
Table~\ref{tab:system} lists the remainder: the 4 departures this paper's
rule does not account for, and then a final row for everything else, which
is the one that matters most. Beyond the departures in those 2 tables, every component is plain
fine-tuning. \textit{The system is plain because the
task justified keeping so little, and knowing what it did justify is what
let us take the rest out.}

\begin{table*}[!t]
\centering
\caption{The departures from ordinary fine-tuning that the chain of
Table~\ref{tab:resource} does not explain. Those 6 run from a piece of task
knowledge through to a component; these 4 rest on weaker grounds, which we state rather than hide. Three of
them were settled by a measurement rather than by a piece of task knowledge;
the first has no support from this audit at all, and we keep it for a reason
of a different kind, given in the text. The final row is the rest of the
system.}
\label{tab:system}
\footnotesize
\setlength{\tabcolsep}{4pt}
\renewcommand{\arraystretch}{1.05}
\begin{tabular}{@{}p{0.265\textwidth}p{0.295\textwidth}p{0.360\textwidth}@{}}
\toprule
In the system & What it does & The comparison behind it \\
\midrule
Three seeds averaged in risk stage 1 & reduces the variance of one training run & No measurable gain on factors; the 1 departure the audit does not justify \\
All 11 factor members kept & the set was fixed before it was scored & Greedy selection: $+0.0063$ against $+0.0157$ \\
Taggers see the predicted level & conditioning, not a model per level & The coupling in Table~\ref{tab:data} \\
7 taggers rather than 8 & one member was making the panel worse & Leave-one-out pruning: $+0.0010$ evidence, unresolved \\
\addlinespace
Everything else & plain fine-tuning, plain heads, plain data & The 26 comparisons that did not survive \\
\bottomrule
\end{tabular}
\end{table*}

\textbf{What the system carries that it should not.} One part of it we
know to be wrong. When we measure how much each factor member contributes
beyond the others, the pool's score peaks at 6 members and declines slowly
thereafter, by per-member amounts that are individually inside the null band and
visible only when the whole curve is drawn, so 5 of the 11 we shipped show no individually resolvable contribution. We
shipped them anyway, and the reason is the same in-sample bias that runs through this
paper. Choosing the 6 would have meant selecting members on the same
held-out predictions used to judge the result, and when we ran that
selection properly it returned $+0.0063$ against the $+0.0157$ of the set
we had fixed in advance. Faced with a subset we believed was better and no
untainted way to identify it, we kept the larger pool and accepted the
cost. It is a small cost and an honest one, and \textit{we record it here because a
system paper that lists only the parts that worked is not describing a
system.}

\textbf{The system stood third of 53 on the public leaderboard, and that is
the least durable thing in this paper.} Table~\ref{tab:leaderboard} gives that
standing. It is the public board at the close of submission, not the
organizers' final evaluation, which is run afterwards on authors withheld from
this set; we report it as the measurement it is. The organizers report risk and
evidence as a single number, 0.8096, which is $(0.4R + 0.3E)/0.7$. We
recovered the parts with probe submissions that held one component fixed
and varied the other, giving risk 0.8203 and evidence 0.7953; those values
reproduce both the joint number and the composite exactly. Both score higher than the factor output at 0.7045, though the 3 metrics are
not on a common scale. The factor output is also where codebook conditioning
helped most.

\begin{table}[t]
\centering
\caption{Public leaderboard standing at the close of submission, on the
blind leaderboard set of 378 posts whose labels we never see, with the 2 entries above and the 2 below ours to
indicate the spacing of the field. The organizers' final evaluation is run
separately, on authors this set does not contain.}
\label{tab:leaderboard}
\setlength{\tabcolsep}{5pt}
\begin{tabular}{@{}lccccc@{}}
\toprule
& First & Second & \textbf{Third (ours)} & Fourth & Fifth \\
\midrule
composite & 0.7951 & 0.7813 & \textbf{0.7781} & 0.7745 & 0.7738 \\
\bottomrule
\end{tabular}
\end{table}

What we would rather be judged on is the set of things that did not have to be there,
and the reason most of them did not survive.

% =====================================================================
\section{Conclusion}

We built a system for reading posts and assessing suicide risk, reached for
the techniques the field agrees on, and found that most of them had nothing
to work on here. \textbf{The finding we would carry forward is not which
techniques transferred, but that the knowledge a task already contains is
enough to decide which components that task needs, and therefore to remove
the ones it does not.} Reading the task first is what let us take the rest
of the machinery out. Of 31 pre-specified comparisons drawn from
7 families of technique, 5 held and 26 did not. \textit{We know of no audit
of this breadth on a task of this shape: each technique has been studied on
its own, elsewhere, but the playbook as a whole has not been put to the
test where practitioners in this area actually work.} Most produced effects
smaller than anything this corpus can resolve, and 6 made a component
measurably worse. The biggest single improvement to the finished
system was not a technique at all. It was a repair: the predictions we
tuned our cutoffs on were not the predictions we finally applied them
to.

What survived was not invented by us. Three of the 5 come straight from the literature: definition conditioning, the
class-balanced loss and ensemble diversity. The other 2, rules scoped to the
evidence clause and the routing of one class, are ordinary moves scoped to
this task. A
technique paid here when the task supplied the particular thing it is built
to consume, and paid nothing when it did not. Written
definitions helped because the categories came from a codebook, and the
identical move failed on the output the codebook does not define. A
class-balanced loss helped because the imbalance it corrects for is real
here. Diverse ensemble members helped because their errors correlate less
with the pool than any same-family member's do. \textbf{None of this can be read off a technique's name or from its
published results.} Several of the strongest design choices follow from
information already written down in the task, chiefly the clinical codebook
and the annotation scheme; the others follow from error and score-distribution properties we had to
measure. \textbf{The first kind is visible on a page of the annotation
guidelines before any model is trained.}
We have called the first of these 2 findings task-conditioned technique
selection and the second deployment-consistent calibration. Together they
are why the submitted system is plainer than the playbook would have made
it, rather than more elaborate.

\textbf{Five principles for building in this regime, each with a
measurement behind it.} These are what we
would carry to the next corpus of this shape, in preference to any technique.

\begin{enumerate}
\item \textbf{Definition conditioning beat model scale.} Telling the model
what a label means moved the factor score from 0.3795 to 0.6172. Doubling the
size of the encoder moved it from 0.6038 to 0.6199, inside the seed range and
therefore nothing we can claim. The cheaper intervention was worth an order of
magnitude more than the expensive one.

\item \textbf{Different outputs supplied different knowledge.} The same
definition conditioning scored 0.4830 against 0.7787 when applied to the
evidence output. The codebook that defines the 24 factors does not define
which words justify a judgment, so on that output there was nothing to supply.

\item \textbf{Complementarity beat standalone quality.} The 2 members that
most improved the factor pool were not its 2 strongest, and a selection rule
based on standalone score would have dropped one of them. Letting a greedy
search pick the members gained 0.0063, whereas the set fixed in advance gained 0.0157.

\item \textbf{Decoding and calibration mattered as much as modeling.}
Correcting where the cutoffs sat was worth more, on an already finished
system, than anything we trained. Fitting those same cutoffs on the posts they
are scored against inflates the result by 0.0684, larger than all but 2 of the
modeling effects we measured.

\item \textbf{A technique fails when the task does not supply what it
consumes.} This is the pattern behind principles 1 to 3, and behind the 26
comparisons that did not hold. Principle 4 is the exception that gives the
paper its second finding: the calibration defect was not a technique at all.
\end{enumerate}

Two habits sit underneath all 5. Measure how far apart 2 identical runs land
before believing any improvement, because on this corpus that distance was
0.095, wider than most of what we set out to detect. And check whether the
numbers you fit your decisions on come from the same kind of model as the
numbers you will finally apply them to. Ours did not, and that single mismatch
cost more than any modeling decision we made.

We know of no comparable audit on a task of this shape, and that absence is
why this is an audit rather than a system description. The techniques we
tested are recommended most confidently for exactly the situation faced
here, and nobody has checked here whether they work. We
would rather this audit were superseded than left as the only data point.
Until it is, the rule we would carry forward is this. \textit{Before adding
a component, name the knowledge, structure or measured failure mode it is meant to
exploit, show that the task supplies enough of it to matter, and only then
ask whether the component earns its place. Where nothing can be
named, the component is complexity the task does not support, and the system
is better without it.}

% =====================================================================
\section{Limitations}

\textbf{First, the baseline moved during the campaign.} Our comparisons
were run one at a time against an incumbent that improved as the work
went on, so a technique tested late faced a stronger baseline than one
tested early. We re-tested 5 failed
techniques against the final system and all 5 still failed, which
bounds the concern but does not remove it: the other failures were not
re-tested, and a complete design would re-run every one of them.

\textbf{Second, the ceiling estimate rests on a non-random sample.} It
uses near-duplicate posts, which are not drawn at random from the corpus
and are concentrated among a few authors. It bounds the attainable score rather than locating
it, and a proper estimate would require re-annotating a random sample with
a second set of clinicians.

\textbf{Third, both seed ranges rest on too few runs.} We have 3 for the
factor component and 2 for the risk component, and one configuration
each. The range between the extremes of a handful of runs understates the
range of the distribution they are drawn from, so both ranges are more likely too narrow than too wide, and both were measured at operating
points below the ones the submitted system reaches. We established no band
for the evidence component at all, and any evidence result sitting near a
boundary should be read as unresolved.

\textbf{Fourth, several techniques were sampled rather than swept.} The
capacity ladder is 3 points, not a curve, so the claim that returns
fell off after the first step describes those 3 points and does not
locate where the falloff begins. The comparison between a fine-tuned large
model and a conditioned encoder is budget-dependent: at 3 training
epochs it favors the encoder by 0.0106 and at 6 it favors the large
model by 0.0399, with the folds still improving when we stopped. We report
that comparison as unsettled.

\textbf{Fifth, one setting was never ablated.} Each post was presented to the model together with a marker of the post's
position in its own author's sequence, and this was on in every risk model we
trained. It is computed within an author and never crosses a split boundary,
so it does not defeat the author-disjoint partitions. It was inherited from an
earlier system on the strength of a prior result, which is precisely the
practice this paper criticizes, and we did not catch it until the audit
was complete.

\textbf{Sixth, 31 comparisons is itself a multiplicity problem.} Running many
comparisons, sweeping candidates within several of them, and judging against
an incumbent that moved during the campaign all create room for a favorable
result to appear by chance. The 80 and 90 percent win-rate thresholds are
conventions we fixed in advance rather than calibrated quantities, and we did
not correct across the family of 31. The protocol of
Section~\ref{sec:protocol} is designed to make this harder, and the direction
of the risk is worth stating: it inflates the 5 positives more than the 26
negatives, since a null verdict is not something a search finds by accident.

\textbf{Seventh, this is one audit of one task.} Its external validity is
bounded on several axes at once. The corpus is English, drawn from one
platform, annotated with one codebook, and scored by one metric; we ran no
prospective evaluation in a clinical workflow, no analysis of performance
across demographic groups, and no replication on an independent dataset. The properties
we argue were decisive, small size, 3 coupled outputs, and a
macro-averaged score over steeply unequal categories, are common in
clinical text but are not universal. We claim the pattern holds for this
shape of task and no further. Our leaderboard placement is exact
arithmetic on one fixed set of 378 posts by 36 authors and comes with no
confidence interval; the gaps between the top few entries are small
relative to what we know about variation across author draws, and we would
not claim the same ordering would survive a different sample of writers.

% =====================================================================
\section{Ethical Considerations}

\textbf{Data.} The corpus consists of public posts annotated by clinicians for risk. We accessed it solely under the terms of the shared task, used it only for that task, and did not share it with anyone outside it. Any public release of our code contains no corpus data. Although the posts are publicly accessible, we do not take public availability to eliminate the ethical obligations associated with reusing individuals' online content in research, particularly where the material is sensitive and potentially identifiable \cite{reagle2022spinning}.
Accordingly, we carried out all processing on hardware under our own control. No post, in whole or in part, was sent to any third-party service or hosted model at any stage of this work, including for annotation, generation, or evaluation; every model referred to in this paper was run locally.

\textbf{Examples.} No text in this paper is quoted from the corpus, and no illustrative example proved necessary. Had one been needed, it would have been written by the authors and labeled as such, following common practice in social media research on mental health, under which verbatim quotation
risks re-identifying an author who did not consent to appear in a paper.

\textbf{Intended use, and what this system is not.} A system of this kind
orders a queue rather than making a clinical judgment, and we should be
careful not to treat that as harmless: deciding who is looked at first is
itself consequential triage. What we studied is ranking as an offline research
problem, and nothing here establishes that the system is safe or effective for
operational use. Its
outputs are not diagnoses; they are not risk assessments in the clinical
sense of that term, and they are not fit to trigger any action concerning
a person without a human reading the post. We report a composite of 0.7781
and would not want that number read as evidence of readiness for
deployment; the risk output is wrong often enough that nothing should follow from it on
its own. The 2 kinds of error are not interchangeable either. A missed
high-risk post and a false alarm carry different costs to different people, and a
single aggregate score establishes an acceptable rate for neither. The argument for producing evidence spans
alongside the risk level is exactly that a reviewer should be able to
check the system against the post rather than defer to the system, and the
organizers of the related CLPsych task are careful to say that such spans
do not by themselves constitute an explanation of risk
\cite{chim2024clpsych}. We make no stronger claim for ours.

\textbf{Foreseeable harms.} Two seem to us most likely. A system of this
kind could be used to surveil rather than to help, applied to people who
have not asked for and would not want the attention; the fact that the
posts are public does not make that acceptable. And a tool that appears to
explain itself invites the reviewer to stop checking, which is a
documented failure mode of automated aids in clinical settings. Both are
reasons to keep such systems inside services people have chosen to use,
with a person in the loop who is accountable for what follows \cite{sarkar2023review}.

\textbf{Annotation.} The judgments this work depends on were made by
clinicians reading distressing material at length. We did not carry out
that annotation; the benefit of it is entirely ours.

% =====================================================================
\bibliographystyle{IEEEtran}
\bibliography{refs}

% Generated by IEEEtran.bst, version: 1.14 (2015/08/26)
\begin{thebibliography}{10}
\providecommand{\url}[1]{#1}
\csname url@samestyle\endcsname
\providecommand{\newblock}{\relax}
\providecommand{\bibinfo}[2]{#2}
\providecommand{\BIBentrySTDinterwordspacing}{\spaceskip=0pt\relax}
\providecommand{\BIBentryALTinterwordstretchfactor}{4}
\providecommand{\BIBentryALTinterwordspacing}{\spaceskip=\fontdimen2\font plus
\BIBentryALTinterwordstretchfactor\fontdimen3\font minus \fontdimen4\font\relax}
\providecommand{\BIBforeignlanguage}[2]{{%
\expandafter\ifx\csname l@#1\endcsname\relax
\typeout{** WARNING: IEEEtran.bst: No hyphenation pattern has been}%
\typeout{** loaded for the language `#1'. Using the pattern for}%
\typeout{** the default language instead.}%
\else
\language=\csname l@#1\endcsname
\fi
#2}}
\providecommand{\BIBdecl}{\relax}
\BIBdecl

\bibitem{who2026suicide}
{World Health Organization}, ``Suicide,'' WHO Fact Sheet, 28 August, 2026, \url{https://www.who.int/news-room/fact-sheets/detail/suicide}.

\bibitem{franklin2017risk}
J.~C. Franklin, J.~D. Ribeiro, K.~R. Fox, K.~H. Bentley, E.~M. Kleiman, X.~Huang, K.~M. Musacchio, A.~C. Jaroszewski, B.~P. Chang, and M.~K. Nock, ``Risk factors for suicidal thoughts and behaviors: A meta-analysis of 50 years of research,'' \emph{Psychological Bulletin}, vol. 143, no.~2, pp. 187--232, 2017.

\bibitem{zirikly2019clpsych}
A.~Zirikly, P.~Resnik, {\"O}.~Uzuner, and K.~Hollingshead, ``{CLPsych} 2019 shared task: Predicting the degree of suicide risk in {R}eddit posts,'' in \emph{Proc. Sixth Workshop on Computational Linguistics and Clinical Psychology}, 2019, pp. 24--33.

\bibitem{sheth2021knowledge}
A.~Sheth, M.~Gaur, K.~Roy, and K.~Faldu, ``Knowledge-intensive language understanding for explainable ai,'' \emph{IEEE Internet Computing}, vol.~25, no.~5, pp. 19--24, 2021.

\bibitem{wei2019eda}
J.~Wei and K.~Zou, ``{EDA}: Easy data augmentation techniques for boosting performance on text classification tasks,'' in \emph{Proc. Conf. Empirical Methods in Natural Language Processing (EMNLP-IJCNLP)}, 2019, pp. 6382--6388.

\bibitem{gururangan2020dapt}
S.~Gururangan, A.~Marasovi{\'c}, S.~Swayamdipta, K.~Lo, I.~Beltagy, D.~Downey, and N.~A. Smith, ``Don't stop pretraining: Adapt language models to domains and tasks,'' in \emph{Proc. 58th Annual Meeting of the Association for Computational Linguistics (ACL)}, 2020, pp. 8342--8360.

\bibitem{yin2019entailment}
W.~Yin, J.~Hay, and D.~Roth, ``Benchmarking zero-shot text classification: Datasets, evaluation and entailment approach,'' in \emph{Proceedings of EMNLP-IJCNLP}, 2019.

\bibitem{lin2017focal}
T.-Y. Lin, P.~Goyal, R.~Girshick, K.~He, and P.~Doll{\'a}r, ``Focal loss for dense object detection,'' in \emph{Proc. IEEE Int. Conf. Computer Vision (ICCV)}, 2017, pp. 2999--3007.

\bibitem{cui2019classbalanced}
Y.~Cui, M.~Jia, T.-Y. Lin, Y.~Song, and S.~Belongie, ``Class-balanced loss based on effective number of samples,'' in \emph{Proceedings of the IEEE/CVF Conference on Computer Vision and Pattern Recognition (CVPR)}, 2019.

\bibitem{lakshminarayanan2017ensembles}
B.~Lakshminarayanan, A.~Pritzel, and C.~Blundell, ``Simple and scalable predictive uncertainty estimation using deep ensembles,'' in \emph{Advances in Neural Information Processing Systems 30 (NIPS)}, 2017, pp. 6402--6413.

\bibitem{lipton2014thresholding}
Z.~C. Lipton, C.~Elkan, and B.~Narayanaswamy, ``Optimal thresholding of classifiers to maximize {F1} measure,'' in \emph{Machine Learning and Knowledge Discovery in Databases (ECML PKDD)}, vol. 8725, 2014, pp. 225--239.

\bibitem{he2019bagoftricks}
T.~He, Z.~Zhang, H.~Zhang, Z.~Zhang, J.~Xie, and M.~Li, ``Bag of tricks for image classification with convolutional neural networks,'' in \emph{Proc. IEEE/CVF Conf. Computer Vision and Pattern Recognition (CVPR)}, 2019, pp. 558--567.

\bibitem{narang2021transformer}
S.~Narang, H.~W. Chung, Y.~Tay, L.~Fedus, T.~Fevry, M.~Matena, K.~Malkan, N.~Fiedel, N.~Shazeer, Z.~Lan, Y.~Zhou, W.~Li, N.~Ding, J.~Marcus, A.~Roberts, and C.~Raffel, ``Do transformer modifications transfer across implementations and applications?'' in \emph{Proc. Conf. Empirical Methods in Natural Language Processing (EMNLP)}, 2021, pp. 5758--5773.

\bibitem{longpre2020augmentation}
S.~Longpre, Y.~Wang, and C.~DuBois, ``How effective is task-agnostic data augmentation for pretrained transformers?'' in \emph{Findings of the Association for Computational Linguistics: EMNLP 2020}, 2020, pp. 4401--4411.

\bibitem{byrd2019importance}
J.~Byrd and Z.~C. Lipton, ``What is the effect of importance weighting in deep learning?'' in \emph{Proc. 36th Int. Conf. Machine Learning (ICML)}, 2019, pp. 872--881.

\bibitem{dacrema2019progress}
M.~{Ferrari Dacrema}, P.~Cremonesi, and D.~Jannach, ``Are we really making much progress? a worrying analysis of recent neural recommendation approaches,'' in \emph{Proc. 13th ACM Conf. Recommender Systems (RecSys)}, 2019, pp. 101--109.

\bibitem{cawley2010selection}
G.~C. Cawley and N.~L.~C. Talbot, ``On over-fitting in model selection and subsequent selection bias in performance evaluation,'' \emph{Journal of Machine Learning Research}, vol.~11, pp. 2079--2107, 2010.

\bibitem{dodge2020finetuning}
J.~Dodge, G.~Ilharco, R.~Schwartz, A.~Farhadi, H.~Hajishirzi, and N.~A. Smith, ``Fine-tuning pretrained language models: Weight initializations, data orders, and early stopping,'' 2020, arXiv:2002.06305.

\bibitem{gaurshethbook}
M.~Gaur and A.~P. Sheth, \emph{Knowledge-Infused Learning: Neurosymbolic {AI} for Explainability, Interpretability, and Safety}.\hskip 1em plus 0.5em minus 0.4em\relax Cambridge University Press, 2025.

\bibitem{gaur2019cssrs}
M.~Gaur, A.~Alambo, J.~P. Sain, U.~Kursuncu, K.~Thirunarayan, R.~Kavuluru, A.~Sheth, R.~Welton, and J.~Pathak, ``Knowledge-aware assessment of severity of suicide risk for early intervention,'' in \emph{The World Wide Web Conference (WWW)}, 2019, pp. 514--525.

\bibitem{gaur2018let}
M.~Gaur, U.~Kursuncu, A.~Alambo, A.~Sheth, R.~Daniulaityte, K.~Thirunarayan, and J.~Pathak, ``" let me tell you about your mental health!" contextualized classification of reddit posts to dsm-5 for web-based intervention,'' in \emph{Proceedings of the 27th ACM international conference on information and knowledge management}, 2018, pp. 753--762.

\bibitem{gaur2021characterization}
M.~Gaur, V.~Aribandi, A.~Alambo, U.~Kursuncu, K.~Thirunarayan, J.~Beich, J.~Pathak, and A.~Sheth, ``Characterization of time-variant and time-invariant assessment of suicidality on reddit using c-ssrs,'' \emph{PloS one}, vol.~16, no.~5, p. e0250448, 2021.

\bibitem{tsakalidis2022overview}
A.~Tsakalidis, J.~Chim, I.~M. Bilal, A.~Zirikly, D.~Atzil-Slonim, F.~Nanni, P.~Resnik, M.~Gaur, K.~Roy, B.~Inkster \emph{et~al.}, ``Overview of the clpsych 2022 shared task: Capturing moments of change in longitudinal user posts,'' in \emph{Proceedings of the Eighth Workshop on Computational Linguistics and Clinical Psychology}, 2022, pp. 184--198.

\bibitem{chim2024clpsych}
J.~Chim, A.~Tsakalidis, D.~Gkoumas, D.~Atzil-Slonim, Y.~Ophir, A.~Zirikly, P.~Resnik, and M.~Liakata, ``Overview of the {CLP}sych 2024 shared task: Leveraging large language models to identify evidence of suicidality risk in online posts,'' in \emph{Proc. 9th Workshop on Computational Linguistics and Clinical Psychology (CLPsych)}, 2024, pp. 177--190.

\bibitem{pfa2025}
J.~Li, X.~Wang, H.~Li, Y.~Yan, H.~V. Leong, L.~Feng, N.~X. Yu, and Q.~Li, ``Protective factor-aware dynamic influence learning for suicide risk prediction on social media,'' \emph{arXiv preprint arXiv:2507.10008}, 2025.

\bibitem{cup2024overview}
J.~Li, Y.~Yan, Z.~Zhang, X.~Wang, H.~V. Leong, N.~X. Yu, and Q.~Li, ``Overview of {IEEE} {BigData} 2024 cup challenges: Suicide ideation detection on social media,'' in \emph{2024 IEEE International Conference on Big Data (BigData)}.\hskip 1em plus 0.5em minus 0.4em\relax IEEE, 2024, pp. 8532--8540.

\bibitem{nguyen2024llm}
V.~Nguyen and C.~Pham, ``Leveraging large language models for suicide detection on social media with limited labels,'' in \emph{2024 IEEE International Conference on Big Data (BigData)}, 2024, arXiv:2410.04501.

\bibitem{ji2022mentalbert}
S.~Ji, T.~Zhang, L.~Ansari, J.~Fu, P.~Tiwari, and E.~Cambria, ``{MentalBERT}: Publicly available pretrained language models for mental healthcare,'' in \emph{Proceedings of the Thirteenth Language Resources and Evaluation Conference (LREC)}, 2022.

\bibitem{8970629}
A.~Sheth, M.~Gaur, U.~Kursuncu, and R.~Wickramarachchi, ``Shades of knowledge-infused learning for enhancing deep learning,'' \emph{IEEE Internet Computing}, vol.~23, no.~6, pp. 54--63, 2019.

\bibitem{sheth2020shades}
------, ``Shades of knowledge-infused learning for enhancing deep learning,'' \emph{IEEE Internet Computing}, vol.~23, no.~6, pp. 54--63, 2020.

\bibitem{dalal2024cross}
S.~Dalal, D.~Tilwani, M.~Gaur, S.~Jain, V.~L. Shalin, and A.~P. Sheth, ``A cross attention approach to diagnostic explainability using clinical practice guidelines for depression,'' \emph{IEEE Journal of Biomedical and Health Informatics}, vol.~29, no.~2, pp. 1333--1342, 2024.

\bibitem{gaur2024crest}
M.~Gaur and A.~Sheth, ``Building trustworthy {NeuroSymbolic} {AI} systems: Consistency, reliability, explainability, and safety,'' \emph{AI Magazine}, vol.~45, no.~1, pp. 139--155, 2024, arXiv:2312.06798.

\bibitem{sinha2026llms}
V.~Sinha, P.~Guttal, P.~D.~R. Katike, V.~Sinha, G.~Ndawula, L.~Yoon, A.~Kleinsmith, and M.~Gaur, ``Where do llms fall short in cbt-guided affective reasoning?'' \emph{arXiv preprint arXiv:2607.02885}, 2026.

\bibitem{mohammadi-etal-2024-welldunn}
\BIBentryALTinterwordspacing
S.~Mohammadi, E.~Raff, J.~Malekar, V.~Palit, F.~Ferraro, and M.~Gaur, ``{W}ell{D}unn: On the robustness and explainability of language models and large language models in identifying wellness dimensions,'' in \emph{Proceedings of the 7th BlackboxNLP Workshop: Analyzing and Interpreting Neural Networks for NLP}, Y.~Belinkov, N.~Kim, J.~Jumelet, H.~Mohebbi, A.~Mueller, and H.~Chen, Eds.\hskip 1em plus 0.5em minus 0.4em\relax Miami, Florida, US: Association for Computational Linguistics, Nov. 2024, pp. 364--388. [Online]. Available: \url{https://aclanthology.org/2024.blackboxnlp-1.23/}
\BIBentrySTDinterwordspacing

\bibitem{lu2022imbalance}
H.~Lu, L.~Ehwerhemuepha, and C.~Rakovski, ``A comparative study on deep learning models for text classification of unstructured medical notes with various levels of class imbalance,'' \emph{BMC Medical Research Methodology}, vol.~22, no.~1, p. 181, 2022.

\bibitem{roy2023process}
K.~Roy, Y.~Zi, M.~Gaur, J.~Malekar, Q.~Zhang, V.~Narayanan, and A.~Sheth, ``Process knowledge-infused learning for clinician-friendly explanations,'' in \emph{Proceedings of the AAAI Symposium Series}, vol.~1, no.~1, 2023, pp. 154--160.

\bibitem{sheth2022process}
A.~Sheth, M.~Gaur, K.~Roy, R.~Venkataraman, and V.~Khandelwal, ``Process knowledge-infused ai: Toward user-level explainability, interpretability, and safety,'' \emph{IEEE Internet Computing}, vol.~26, no.~5, pp. 76--84, 2022.

\bibitem{ji2021magicbert}
S.~Ji, M.~H{\"o}ltt{\"a}, and P.~Marttinen, ``Does the magic of {BERT} apply to medical code assignment? a quantitative study,'' \emph{Computers in Biology and Medicine}, vol. 139, p. 104998, 2021.

\bibitem{melis2018sota}
G.~Melis, C.~Dyer, and P.~Blunsom, ``On the state of the art of evaluation in neural language models,'' in \emph{International Conference on Learning Representations (ICLR)}, 2018.

\bibitem{ruffinelli2020olddog}
D.~Ruffinelli, S.~Broscheit, and R.~Gemulla, ``You {CAN} teach an old dog new tricks! on training knowledge graph embeddings,'' in \emph{International Conference on Learning Representations (ICLR)}, 2020.

\bibitem{musgrave2020metric}
K.~Musgrave, S.~Belongie, and S.-N. Lim, ``A metric learning reality check,'' in \emph{Proc. European Conf. Computer Vision (ECCV)}, vol. 12370.\hskip 1em plus 0.5em minus 0.4em\relax Springer, 2020, pp. 681--699.

\bibitem{zheng2022fewnlu}
Y.~Zheng, J.~Zhou, Y.~Qian, M.~Ding, C.~Liao, L.~Jian, R.~Salakhutdinov, J.~Tang, S.~Ruder, and Z.~Yang, ``{FewNLU}: Benchmarking state-of-the-art methods for few-shot natural language understanding,'' in \emph{Proc. 60th Annual Meeting of the Association for Computational Linguistics (ACL)}, 2022.

\bibitem{zhang2021fewsample}
T.~Zhang, F.~Wu, A.~Katiyar, K.~Q. Weinberger, and Y.~Artzi, ``Revisiting few-sample {BERT} fine-tuning,'' in \emph{International Conference on Learning Representations (ICLR)}, 2021.

\bibitem{gorman2019splits}
K.~Gorman and S.~Bedrick, ``We need to talk about standard splits,'' in \emph{Proc. 57th Annual Meeting of the Association for Computational Linguistics (ACL)}, 2019, pp. 2786--2791.

\bibitem{card2020power}
D.~Card, P.~Henderson, U.~Khandelwal, R.~Jia, K.~Mahowald, and D.~Jurafsky, ``With little power comes great responsibility,'' in \emph{Proc. Conf. Empirical Methods in Natural Language Processing (EMNLP)}, 2020, pp. 9263--9274.

\bibitem{reimers2017seeds}
N.~Reimers and I.~Gurevych, ``Reporting score distributions makes a difference: Performance study of {LSTM}-networks for sequence tagging,'' in \emph{Proc. Conf. Empirical Methods in Natural Language Processing (EMNLP)}, 2017, pp. 338--348.

\bibitem{oliver2018ssl}
A.~Oliver, A.~Odena, C.~Raffel, E.~D. Cubuk, and I.~J. Goodfellow, ``Realistic evaluation of deep semi-supervised learning algorithms,'' in \emph{Advances in Neural Information Processing Systems (NeurIPS)}, 2018.

\bibitem{christodoulou2019ml}
E.~Christodoulou, J.~Ma, G.~S. Collins, E.~W. Steyerberg, J.~Y. Verbakel, and B.~Van~Calster, ``A systematic review shows no performance benefit of machine learning over logistic regression for clinical prediction models,'' \emph{Journal of Clinical Epidemiology}, vol. 110, pp. 12--22, 2019.

\bibitem{alambo2019question}
A.~Alambo, M.~Gaur, U.~Lokala, U.~Kursuncu, K.~Thirunarayan, A.~Gyrard, A.~Sheth, R.~S. Welton, and J.~Pathak, ``Question answering for suicide risk assessment using reddit,'' in \emph{2019 IEEE 13th International Conference on Semantic Computing (ICSC)}.\hskip 1em plus 0.5em minus 0.4em\relax IEEE, 2019, pp. 468--473.

\bibitem{kermani2025clpsych}
A.~Kermani, V.~P{\'e}rez-Rosas, and V.~Metsis, ``A systematic evaluation of {LLM} strategies for mental health text analysis: Fine-tuning vs. prompt engineering vs. {RAG},'' in \emph{Proc. 10th Workshop on Computational Linguistics and Clinical Psychology (CLPsych)}, 2025, pp. 172--180.

\bibitem{laurer2024nli}
M.~Laurer, W.~van Atteveldt, A.~Casas, and K.~Welbers, ``Less annotating, more classifying: Addressing the data scarcity issue of supervised machine learning with deep transfer learning and {BERT-NLI},'' \emph{Political Analysis}, vol.~32, no.~1, pp. 84--100, 2024.

\bibitem{du2021selftraining}
J.~Du, E.~Grave, B.~Gunel, V.~Chaudhary, O.~Celebi, M.~Auli, V.~Stoyanov, and A.~Conneau, ``Self-training improves pre-training for natural language understanding,'' in \emph{Proc. Conf. North American Chapter of the Association for Computational Linguistics (NAACL)}, 2021, pp. 5408--5418, arXiv:2010.02194.

\bibitem{shi2023corn}
X.~Shi, W.~Cao, and S.~Raschka, ``Deep neural networks for rank-consistent ordinal regression based on conditional probabilities,'' \emph{Pattern Analysis and Applications}, vol.~26, no.~3, pp. 941--955, 2023.

\bibitem{he2023debertav3}
P.~He, J.~Gao, and W.~Chen, ``{DeBERTaV3}: Improving {DeBERTa} using {ELECTRA}-style pre-training with gradient-disentangled embedding sharing,'' in \emph{International Conference on Learning Representations (ICLR)}, 2023.

\bibitem{qwen25}
{Qwen}, ``{Qwen2.5} technical report,'' \emph{arXiv preprint arXiv:2412.15115}, 2024.

\bibitem{collell2018threshold}
G.~Collell, D.~Prelec, and K.~R. Patil, ``A simple plug-in bagging ensemble based on threshold-moving for classifying binary and multiclass imbalanced data,'' \emph{Neurocomputing}, vol. 275, pp. 330--340, 2018, arXiv:1606.08698.

\bibitem{reagle2022spinning}
J.~Reagle and M.~Gaur, ``Spinning words as disguise: Shady services for ethical research?'' \emph{First Monday}, 2022.

\bibitem{sarkar2023review}
S.~Sarkar, M.~Gaur, L.~K. Chen, M.~Garg, and B.~Srivastava, ``A review of the explainability and safety of conversational agents for mental health to identify avenues for improvement,'' \emph{Frontiers in Artificial Intelligence}, vol.~6, p. 1229805, 2023.

\end{thebibliography}

\end{document}